\documentclass[letterpaper, 10 pt, conference]{ieeeconf}  

\IEEEoverridecommandlockouts                              

\usepackage{graphics} 
\usepackage{epsfig} 
\usepackage{amsmath} 
\usepackage{amssymb}  
\usepackage{dsfont}
\usepackage{cite}
\usepackage{array}
\usepackage{multirow}
\usepackage{hyperref}
\usepackage{subcaption}
\usepackage{caption}
\usepackage{tikz}
\DeclareCaptionLabelSeparator{periodspace}{.\quad}
\DeclareMathOperator*{\argmin}{arg\,min}

\title{\LARGE \bf
   3DGS-CD: 3D Gaussian Splatting-based Change Detection for Physical Object Rearrangement
}

\author{Ziqi Lu$^{1}$, Jianbo Ye$^{2}$, and John Leonard$^{1}$
\thanks{*This work was supported by ONR Neuro-Autonomy MURI grant N00014-19-1-2571, ONR DURIP grant N00014-23-12164, and the MIT Portugal Program}
\thanks{$^{1}$Ziqi Lu and John Leonard are with the Computer Science and Artificial Intelligence Laboratory,
        Massachusetts Institute of Technology, Cambridge, MA 02139, USA
        {\tt\small ziqilu, jleonard@mit.edu}}%
\thanks{$^{2}$Jianbo Ye
        {\tt\small jianboye.ai@gmail.com}}%
}

\begin{document}

\maketitle
\thispagestyle{empty}
\pagestyle{empty}

\begin{abstract}
We present 3DGS-CD, the first 3D Gaussian Splatting (3DGS)-based method for detecting physical object rearrangements in 3D scenes.
Our approach estimates 3D object-level changes by comparing two sets of unaligned images taken at different times.
Leveraging 3DGS's novel view rendering and EfficientSAM's zero-shot segmentation capabilities,
we detect 2D object-level changes, which are then associated and fused across views to estimate 3D change masks and object transformations.
Our method can \textit{accurately} identify changes in \textit{cluttered environments} using \textit{sparse} (\textit{as few as one}) post-change images
within as little as \textit{18s}.
It does not rely on depth input, user instructions, pre-defined object classes, or object models~-- 
An object is recognized simply if it has been re-arranged.
Our approach is evaluated on both public and self-collected real-world datasets,
achieving up to \textit{14\% higher accuracy} and \textit{three orders of magnitude faster} performance compared to the state-of-the-art radiance-field-based change detection method.
This significant performance boost enables a broad range of downstream applications,
where we highlight three \textit{key use cases}: object reconstruction, robot workspace reset, and 3DGS model update.
Our code and data will be made available at \url{https://github.com/520xyxyzq/3DGS-CD}.
\end{abstract}

\section{INTRODUCTION}


3D change detection involves identifying changed objects or regions in the environment from two sets of local observations taken at different times or from new observations of a previously modeled scene.
It is a critical task in robotics as it can accommodate not only short-term, continuously observed scene dynamics but also long-term scene changes where the transition process is typically unobserved.
Capturing such long-term changes is particularly important for robot operations,
as environmental changes often occur without being noticed or monitored.

Despite the success of 3D change detection with depth input using scene representations such as TSDF \cite{fehr2017tsdf}, 3D point cloud \cite{finman2013toward} and neural descriptor fields \cite{fu2022robust},
detecting changes from multi-view RGB images remains a challenging problem.
Traditional methods have relied on hand-crafted techniques like voxelization \cite{pollard2007change}, multi-view stereo \cite{sakurada2013detecting}, and image warping \cite{palazzolo2018fast} to identify changes in unaligned images and lift to 3D.
These approaches are particularly sensitive to occlusions and lighting variations,
especially when there are large viewpoint differences between the two sets of images.

The emergence of radiance field models,
such as neural radiance fields (NeRF) \cite{mildenhall2021nerf} and 3DGS \cite{kerbl20233d},
presents new opportunities to address these challenges.
These models provide high-fidelity representations of scene geometry and appearance,
with novel-view rendering capabilities that enable the generation of photo-realistic images and dense depths at arbitrary viewpoints.
This allows for direct comparison of pre- and post-change images from the same viewpoint.

NeRF-based solutions have been explored to a limited extent~\cite{lu2024fast,huang2023c}.
However, these methods are constrained by the computational cost of NeRF's ray-casting-based rendering.
In contrast, 3DGS offers a more efficient alternative,
achieving real-time rendering with comparable or even superior quality.

In our work, we leverage 3DGS as scene representation to identify 3D scene changes including object removal, insertion and movement from multi-view images.
We exploit the zero-shot segmentation capability of EfficientSAM~\cite{xiong2024efficientsam} to compare pre- and post-change images at the same viewpoints,
associating and fusing the 2D object changes to obtain accurate object 3D change masks and 6D pose changes.

Our method has the following key advantages:
(1)~It can handle \textbf{sparse} post-change image inputs, requiring as few as \textit{a single new image} to detect 3D changes.
(2)~It requires no depth sensors or monocular depth estimators.
(3)~It does not rely on pre-defined object classes, models or object detectors
-- An object is recognized simply if it has been moved, removed or inserted.
(4)~It requires no user instructions such as user-provided click or language prompts.
(5)~It is agnostic to the choice of radiance field models and could, in principle, work with any, such as NeRF\cite{mildenhall2021nerf} and 2DGS~\cite{huang20242d}.

Our method is evaluated on both public and self-collected real-world datasets,
achieving up to \textbf{14\% higher accuracy} and \textbf{three magnitudes faster} performance compared to the state-of-the-art NeRF-based method.
This significant performance improvement enables a wide range of real-world applications, including:
(1)~object removal as a prompt for object reconstruction (Sec.~\ref{sec:recon});
(2)~robot workspace reset (Sec.~\ref{sec:reset});
(3)~3DGS model update (Sec.~\ref{sec:update}).

\begin{figure*}[htb!]
    \centering
    \includegraphics[width=\linewidth]{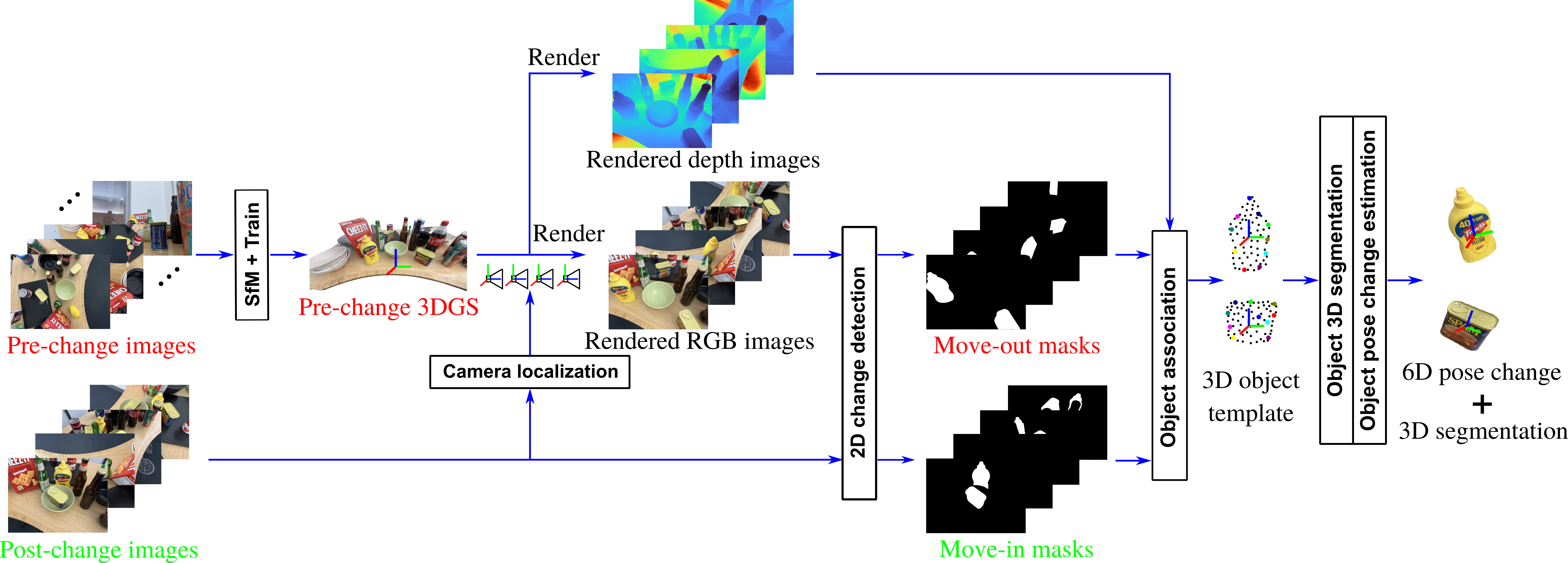}
    \caption{
        Our method detects 3D object-level changes from pre- and potentially sparse post-change images of a 3D scene.
        We first train a 3DGS model on pre-change images (Sec.\ref{sec:3dgs}),
        localize the post-change cameras with respect to this model (Sec.\ref{sec:loc}),
        and render RGB-D images at post-change views for 2D change detection using EfficientSAM \cite{xiong2024efficientsam} (Sec.\ref{sec:2d}).
        The detected 2D object segments are associated across the post-change views (Sec.~\ref{sec:ass}) to initialize 3D object templates.
        These templates are used to classify object change types and query EfficientSAM on pre-change views to obtain additional 2D object segments,
        which are fused to obtain 3D object segments (Sec.~\ref{sec:seg}).
        For moved objects, we leverage image-template feature correspondences to estimate their 6D pose changes  (Sec.~\ref{sec:pose_est})
        and refine the estimates by a render-and-compare approach (Sec.~\ref{sec:global}).
    }
    \label{fig:method}
\end{figure*}

\section{Related work}


\subsection{Change detection from RGB images}
2D change detection from image pairs has been a long-studied problem.
Most existing methods assume minimal viewpoint differences between images \cite{alcantarilla2018street,zheng2021change,bandara2022transformer,codegoni2023tinycd,li2024new,li2024open, wang2024mtp}.
While some approaches can handle larger viewpoint shifts \cite{jst2015change,sachdeva2023changeold, sachdeva2023change},
they are often restricted to the provided views and cannot fully capture physical changes in 3D.

Traditionally, 3D change detection from multi-view RGB images has relied on hand-crafted techniques to identify changes from unaligned images~\cite{pollard2007change, sakurada2013detecting, palazzolo2018fast}.
NeRF-based methods have been explored to directly compare pre- and post-change images from the same viewpoints, but only to a limited extent.
NeRF-Update \cite{lu2024fast} combines NeRF with the Segment Anything Model (SAM) \cite{kirillov2023segment} to detect object movements,
while C-NeRF \cite{huang2023c} models scene changes as direction-consistent radiance differences between two NeRFs.
However, these methods are limited by NeRF's slow rendering speed.
In contrast, 3DGS offers real-time rendering speeds through its rasterization-based rendering pipeline.
While early works such as SplatPose \& Detect~\cite{kruse2024splatpose} have explored the use of 3DGS for anomaly detection,
there has been little effort to directly apply 3DGS to the task of change detection.
In this work, we leverage 3DGS's efficient novel-view rendering capability for fast 3D scene change detection.
\subsection{3DGS as scene representation}
3DGS has recently emerged as a key scene representation in robotics 
but it lacks native support for change representation.
While dynamic 3DGS methods~\cite{luiten2023dynamic,yang2023real,lin2024gaussian,gao2024gaussianflow, wu20244d,guo2024motion,sun20243dgstream, zhang2024monst3r} can capture short-term dynamics in video sequences,
many of them typically rely on continuous spatio-temporal observation of the transformation process.
Our method addresses this gap by detecting changes in 3DGS-represented scenes,
offering a more robust solution for capturing 3D scene dynamics.

\section{Background}
\subsection{3D Gaussian Splatting (3DGS)}

3DGS \cite{kerbl20233d} uses a collection of 3D Gaussians $\mathcal{G}$ to explicitly encode the geometry and appearance of a 3D scene.
Each 3D Gaussian is defined by its mean (position) $\mu\in\mathbb{R}^3$,
covariance matrix $\Sigma\in\mathbb{R}^{3\times 3}$,
opacity $\alpha\in\mathbb{R}$ and color $c\in\mathbb{R}^3$.
To render an image from a given view with known camera parameters,
the 3D Gaussians are projected (i.e. splatted) onto the camera's 2D image plane.
For each pixel, the subset of 3D Gaussians $\mathcal{G}_p$ that project onto that pixel are gathered and sorted by depth.

The color of a pixel $p$ can be computed from the splatted Gaussians using alpha-compositing as:
\begin{align}\label{eqn:3dgs}
    C(p)=\sum_{i\in\mathcal{G}_p} c_i \alpha_i(p) \prod_{j=1}^{i-1}\left(1-\alpha_j(p)\right)
\end{align}
where $c_i$ is the color of the $i$-th Gaussian
and $\alpha_i(p)$ is the opacity of the $i$-th splatted Gaussian evaluated at pixel $p$.

The depth for pixel $p$ can be similarly computed by:
\begin{align}\label{eqn:3dgs_depth}
    D(p)=\sum_{i\in\mathcal{G}_p} d_i \alpha_i(p) \prod_{j=1}^{i-1}\left(1-\alpha_j(p)\right)
\end{align}
where $d_i$ is the depth of the $i$-th splatted Gaussian.

During 3DGS training, 
the Gaussian parameters are optimized by minimizing the error between the rendered and ground truth images over a set of training views.
The optimization uses a loss function that combines pixel-wise L1 loss with the structural loss D-SSIM.

\subsection{EfficientSAM}

The segment anything model (SAM) \cite{kirillov2023segment} is a foundation model for image segmentation from input prompts such as boxes or points.
EfficientSAM\cite{xiong2024efficientsam} is a light-weight SAM model enabled by SAM-leveraged masked image pretraining.
EfficientSAM adopts an encoder-decoder architecture, i.e. $\mathcal{S}=\mathcal{S}_e \circ \mathcal{S}_d$.
The encoder $\mathcal{S}_e$ predicts the embedding $f\in\mathbb{R}^{h\times w\times d}$ for input image $I\in\mathbb{R}^{H\times W\times3}$,
and the decoder $S_d$ takes the image embedding $f$ and the prompt $\mathcal{P}$ to output the segmentation mask and the prediction confidence.
\begin{align}
    \mathbf{M}, \mathbf{C} = \mathcal{S}(I, \mathcal{P}) = \mathcal{S}_d\left(\mathcal{S}_e(I), \mathcal{P}\right)
\end{align}

\begin{figure*}[htb!]
    \includegraphics[width=\linewidth]{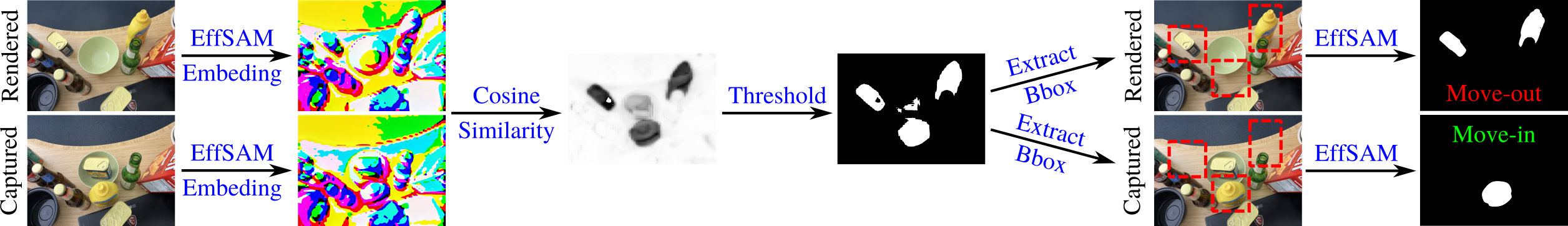}
    \caption{
        EfficientSAM embedding for 2D change detection at a post-change view (Sec.~\ref{sec:2d}).
        While the 2D object masks may be incomplete or missing due to occlusions or prediction failures,
        associating and fusing masks from multiple views (Sec.~\ref{sec:ass}) help recover more complete object templates.
        }
    \label{fig:2dcd}
\end{figure*}

In our work,
we use the EfficientSAM image embedding $f=\mathcal{S}_e(I)$ for 2D object-level change detection,
and we use box prompts $\mathcal{B}$ to query EfficientSAM to obtain 2D object segments and confidence,
i.e. $\mathbf{M}, \mathbf{C} = \mathcal{S}(I, \mathcal{B})$.

\section{Methodology}
Our method aims to detect object-level changes in a 3D scene from two sets of RGB images:
(1)~Pre-change images $\{I_m\}_{m=1}^M$ capturing the initial static state of the scene.
(2)~Potentially sparse post-change images $\{I^\prime_n\}_{n=1}^N$ observing the changed state of the scene.
Our method outputs
(1) 3D segmentation $\mathcal{M}: \mathbb{R}^3\to\{0,1\}$ and
(2) pose change $\mathcal{T}\in\text{SE}(3)$ for each re-arranged object.

The pipeline of our method is developed to be modular.
As shown in Fig.~\ref{fig:method}, it consists of the following submodules:
(1) Pre-change 3DGS training (Sec.~\ref{sec:3dgs});
(2) post-change camera localization (Sec.~\ref{sec:loc});
(3) 2D change detection on post-change views (Sec.~\ref{sec:2d});
(4) Object association across post-change views (Sec.~\ref{sec:ass});
(5) Object 3D  segmentation (Sec.~\ref{sec:seg});
(6) Object pose change estimation (Sec.~\ref{sec:pose_est}) and refinement (Sec.~\ref{sec:global}).
Beyond that, we also discuss how to project the 3D object segmentations to novel views with occlusion handling (Sec.~\ref{sec:proj})
and provide implemetation details in Sec.~\ref{sec:impl}.

\subsection{Pre-change 3DGS training}\label{sec:3dgs}
We first train a 3DGS model $\mathcal{G}$ on the pre-change images $\{I_m\}_{m=1}^M$.
Following the standard 3DGS training procedure~\cite{kerbl20233d}, 
we use a structure-from-motion (SfM) algorithm~\cite{sarlin2019coarse} to estimate pre-change camera parameters and generate a sparse point cloud,
initializing 3D Gaussians for subsequent optimization (i.e. training).
\footnote{
    For the success of SfM, 
    we typically need the pre-change images to sufficiently cover the pre-change scene.
    But this requirement can be relaxed with sparse-view 3DGS training techniques, e.g. InstantSplat \cite{fan2024instantsplat}.
}
\subsection{Post-change camera localization}\label{sec:loc}
Within the coordinate frame of the pre-change 3DGS, 
we estimate the camera poses $\{\mathcal{T}_{c^{\prime}_n}\in\text{SE}(3)\}_{n=1}^N$ of the post-change images $\{I^\prime_n\}_{n=1}^N$.
Assuming most of the scene observed by the post-change images remains unchanged, 
we use the standard visual localization procedure~\cite{sarlin2019coarse} to localize the post-change cameras against the SfM point cloud.
These initial pose estimates will be further refined in Sec.~\ref{sec:global}. 

\subsection{EfficientSAM embedding for 2D change detection}\label{sec:2d}
At the post-change camera views,
we use the pre-change 3DGS to render RGB images
and compare them with the captured post-change images to detect object-level changes.

We first coarsely detect 2D changes with EfficientSAM~\cite{xiong2024efficientsam} image embeddings (Fig.~\ref{fig:2dcd}).
The SAM \cite{kirillov2023segment} image embedding,
which encodes rich semantic information, 
has been shown to enable robust zero-shot change detection \cite{lu2024fast,ding2024adapting}.
However, probing SAM's latent space can be computationally expensive.
We therefore adopt the lightweight EfficientSAM encoder to extract image embeddings on the rendered and captured post-change images, 
and threshold their cosine similarity maps to obtain the change masks:
\begin{align}\label{eqn:change_map}
    \centering
    \mathbf{M}^{\text{change}}_n = \left<\mathcal{S}_e\left[\mathcal{R}(\mathcal{T}_{c_n^\prime}; \mathcal{G})\right], \mathcal{S}_e(I^\prime_n)\right> > \text{thresh.}
\end{align}
Here, $\mathcal{R}(\mathcal{T}_{c_n^\prime}; \mathcal{G})$ denotes the rendered image at view $n$,
the image embeddings are upscaled to the original image resolution before the differencing,
and the threshold is dynamically set by Otsu's method~\cite{otsu1975threshold}.

To mitigate the influence of image misalignment and 3DGS floating artifacts on the detection results,
we follow \cite{lu2024fast} to pre-align and blur the rendered and captured images.

From the change masks, we extract contours occupying substantial 2D areas
and use their enclosing bounding boxes to query EfficientSAM on the rendered and captured images:
\begin{align}
\centering
\mathbf{M}_n, \mathbf{C}_n & = \mathcal{S}\left[\mathcal{R}(\mathcal{T}_{c_n^\prime}), \mathcal{B}(\mathbf{M}^{\text{change}}_n)\right] \nonumber\\
\mathbf{M}^{\prime}_n, \mathbf{C}^\prime_n & = \mathcal{S}\left[I^\prime_n, \mathcal{B}(\mathbf{M}^{\text{change}}_n)\right]
\end{align}
As shown in Fig.~\ref{fig:2dcd},
the high-confidence 2D segments on the rendered images are used as move-out masks (previous location) for the re-arranged objects,
whereas the high-confidence masks on the captured images are identified as object move-in masks (new location):
\begin{align}\label{eqn:conf_thresh}
    \{\mathrm{M}_{n,l}\}_{l=1}^L = \{\mathbf{M}_{n}\mid \mathbf{C}_{n} >0.95\} \nonumber\\
    \{\mathrm{M}_{n,l}^\prime\}_{l=1}^{L^\prime} = \{\mathbf{M}^\prime_{n}\mid \mathbf{C}_{n}^\prime >0.95\}
\end{align}
where $l$ indexes the high-confidence masks for view $n$.

\subsection{Change-aware object association}\label{sec:ass}
We associate the 2D object segments across post-change camera views to initialize the re-arranged objects in 3D.
As Fig.~\ref{fig:method} shows,
each initialized object template consists of a dense point cloud and a sparse visual feature point cloud.
These templates are built incrementally by aggregating spatially proximate and semantically similar object segments across views.

We first use the pre-change 3DGS to render depths within the object move-out masks and back-project them to 3D to get per-view partial object point clouds:
\begin{align}
    X_{n,l} = \pi^{-1}\left[\mathcal{R}^d(\mathcal{T}_{c_n^\prime}) \cdot \mathrm{M}_{n,l}\right]
\end{align}
where $\pi^{-1}(\cdot)$ represents the back-projection function
and $\mathcal{R}^d(\cdot)$ denotes the 3DGS depth rendering function as in Eq.~\ref{eqn:3dgs_depth}.

We also extract the EfficientSAM embeddings for the object move-out masks 
and compute their L1-median to obtain per-view per-object representative embedding vectors:
\begin{align}
    f_{n,l} = \text{med}\left[S_e(\mathcal{R}(\mathcal{T}_{c_n^\prime})) \cdot \mathrm{M}_{n,l}\right]
\end{align}
These embedding vectors are only used during object association and will not be included in the object templates.

We further detect sparse visual features (e.g. SuperPoint \cite{detone2018superpoint}) within the object move-out masks and back-project them to 3D to form the per-view sparse feature point clouds: 
\begin{align}
    F_{n,l} = \pi^{-1}\left\{\phi\left[\mathcal{R}(\mathcal{T}_{c_n^\prime}) \cdot \mathrm{M}_{n,l}\right]\right\}
\end{align}
where $\phi$ is the feature detection function, and $\pi^{-1}$ back-projects the 2D feature locations to 3D. 

Starting from the object move-out masks on the first view,
we iteratively associate the segments across remaining views by matching the partial point clouds.
Specifically, we use the Hungarian algorithm to perform nearest-neighbor matching between the current object point cloud and the next segment with the Hausdorff distance.
The matched new segments are added to the object templates while the unmatched segments are used to initialize new object templates.
We also verify the semantic similarity of the matched segments by comparing their EfficientSAM embeddings.
If their cosine similarity is below 40\%,
a new object template is created instead for the new segment.
After the association is done, we obtain the object templates as the union of multi-view point clouds and features:
\begin{align}
    \{ O_k = \{X_k, F_k\} = \cup_n \{X_{n,k}, F_{n,k}\} \}_{k=1}^K
\end{align}
where $k$ indexes the high-confidence masks across post-change views that associates to the $k$-th object.

Note that associating object move-out masks only initializes objects at their pre-change locations.
We also need to associate the object templates with the object move-in masks to identify the object re-arrangement.
We similarly extract the multi-view EfficientSAM embeddings and visual features within the object move-in masks:
\begin{align}
    f_{n,l}^\prime &= \text{med}\left[S_e(I^\prime_n) \cdot \mathrm{M}_{n,l}^\prime\right] \nonumber\\
    F_{n,l}^\prime &= \phi\left(I^\prime_n \cdot \mathrm{M}_{n,l}^\prime\right)
\end{align}

We match the 2D object features $F^\prime_{n,l}$ on post-change images with 3D template features $F_{k}$ to establish 2D-3D correspondences.
Based on the matching results, we can categorize the object-level changes as:
\begin{itemize}
    \item \textbf{Moved object}: object template with sufficient inlier 2D-3D correspondences
    \item \textbf{Removed object}: object template without sufficient inlier 2D-3D correspondences
    \item \textbf{Inserted object}: post-change object segments without sufficient inlier matches to any object template
\end{itemize}
The inlier matches are identified with the RANSAC-PnP algorithm (Sec.~\ref{sec:pose_est}).

For ``inserted" post-change object segments,
we adopt the same object association procedure as above to initialize their object templates. 
Since depths are unavailable for post-change views,
we match the segments using cosine similarity of the embedding vectors $f_{n,l}^\prime$.

\subsection{Object 3D segmentation}\label{sec:seg}
We perform multi-view mask fusion to establish the 3D segmentation mask $\mathcal{M}_k$ for each object template $O_k$.

Since the move-out masks $\mathrm{M}_{n,l}$ on potentially sparse post-change views may not be sufficient for accurate fusion,
we project the point clouds of \textit{moved} and \textit{removed} objects\footnote{
    This step is skipped for \textit{inserted objects} since they don't have move-out masks.
    We therefore require more post-change images to more accurately segment inserted objects.    
}
onto pre-change images and query EfficientSAM with their 2D bounding boxes for additional object move-out masks.
To ensure robustness against occlusion, we only fuse non- or slightly-occluded object masks,
filtering out those that enclose less than 80\% of the projected 2D object points.

For multi-view mask fusion,
we initialize a 3D binary voxel grid $\mathcal{M}_k\in\{0,1\}^{N_x\times N_y\times N_z}$ around each object template.
A given voxel $x$ is considered inside the object if it projects into the majority of the object's 2D masks:
\begin{align}\label{eqn:seg}
    \mathcal{M}_k(x) = 
    \frac{1}{N_k} \sum_{n_k=1}^{N_k} \mathds{1}_{\left\{\mathrm{M}_{n_k,k}\left[\pi(x)\right] = 1 \right\}} > 0.95
\end{align}
where $\pi$ is the projection function and $\mathds{1}$ the indicator function.
$n_k$ indexes all pre- and post-change images that has high-confidence low-occlusion masks for the $k$-th object.
At runtime, the object voxel grids can be queried quickly with arbitrary 3D positions using trillinear interpolation,
and a point is considered inside the object if its occupancy probability is above $0.5$.

\subsection{Object pose change estimation}\label{sec:pose_est}
For \textit{moved objects}, we use the 2D-3D correspondences from Sec.\ref{sec:ass},
and apply the standard RANSAC-PnP algorithm to coarsely estimate the object pose changes.
The initial estimates are further optimized in Sec.~\ref{sec:global}.

\begin{figure*}[htb!]
    \centering
    \begin{subfigure}[b]{0.325\linewidth}
        \includegraphics[width=0.49\linewidth]{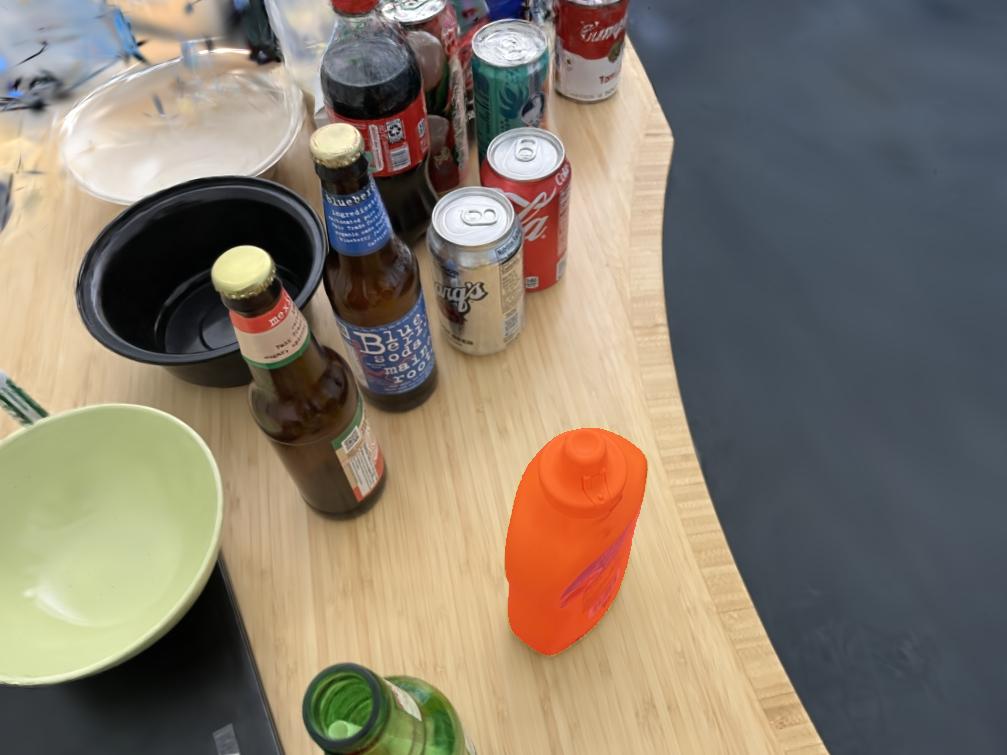}
        \includegraphics[width=0.49\linewidth]{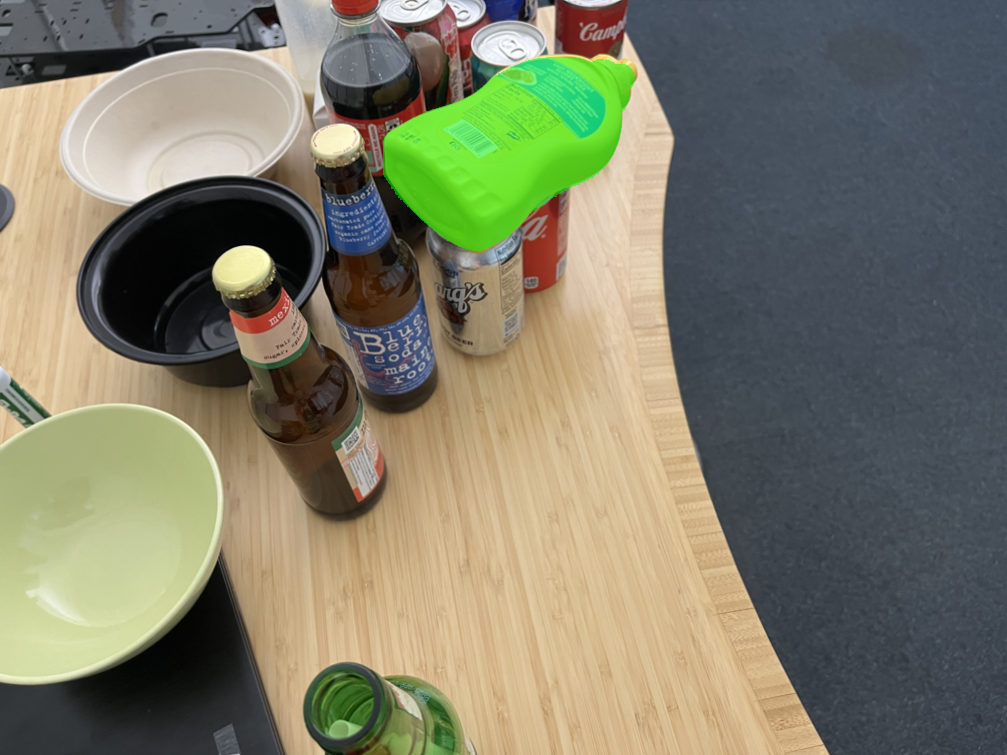}
        \caption{Mustard}
    \end{subfigure}
    \begin{subfigure}[b]{0.325\linewidth}
        \includegraphics[width=0.49\linewidth]{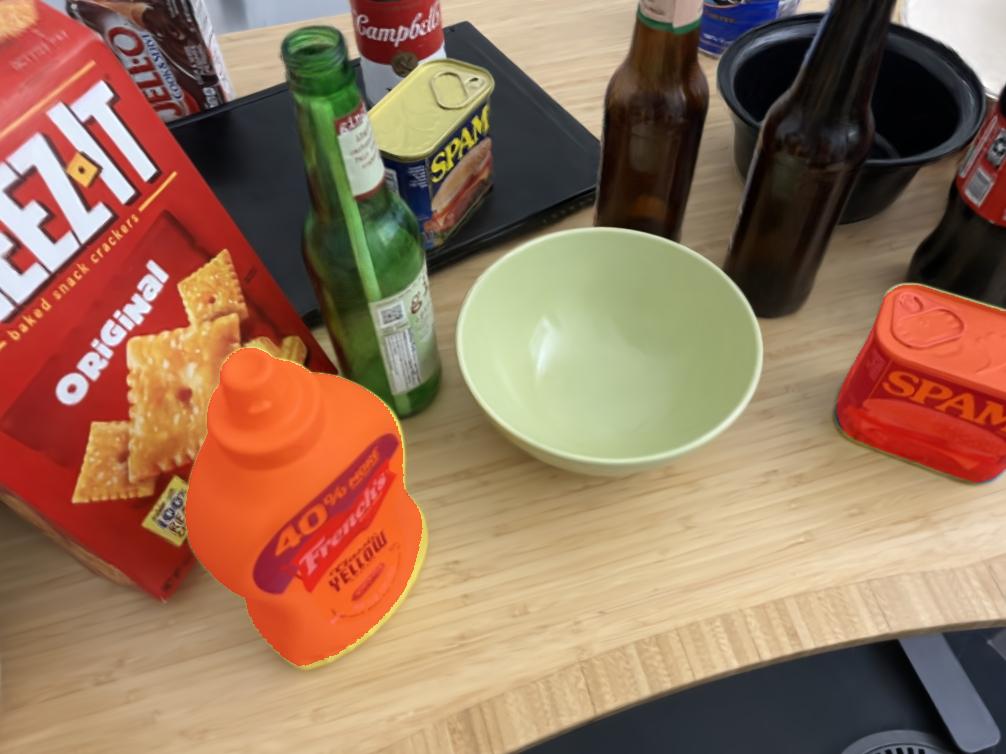}
        \includegraphics[width=0.49\linewidth]{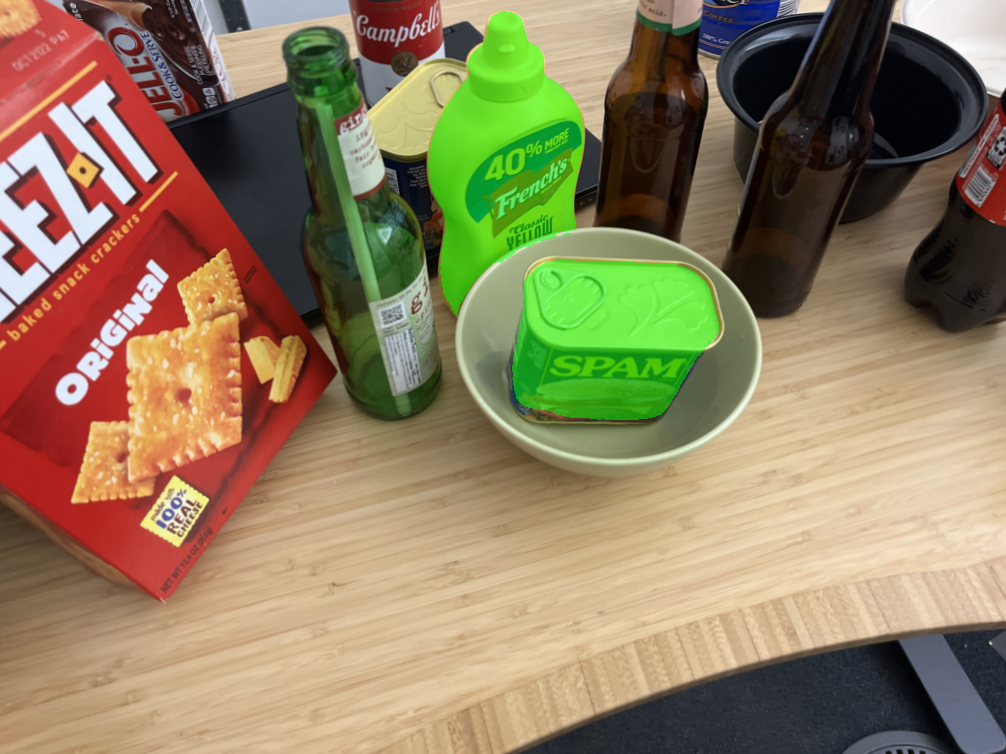}
        \caption{Desk}
    \end{subfigure}
    \begin{subfigure}[b]{0.325\linewidth}
        \includegraphics[width=0.49\linewidth]{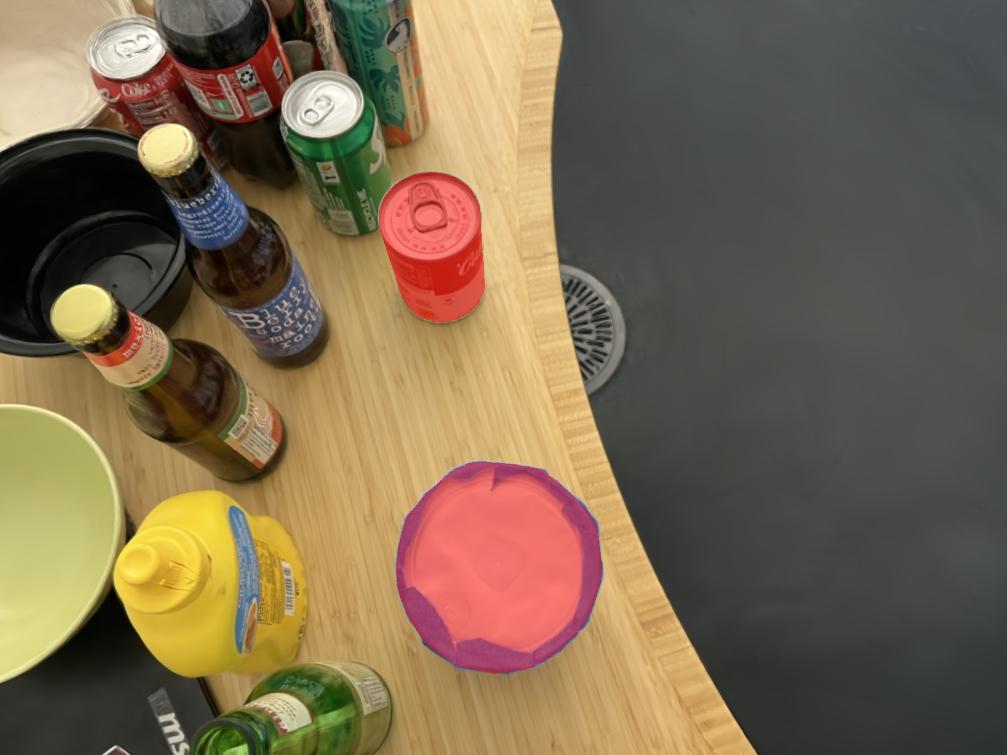}
        \includegraphics[width=0.49\linewidth]{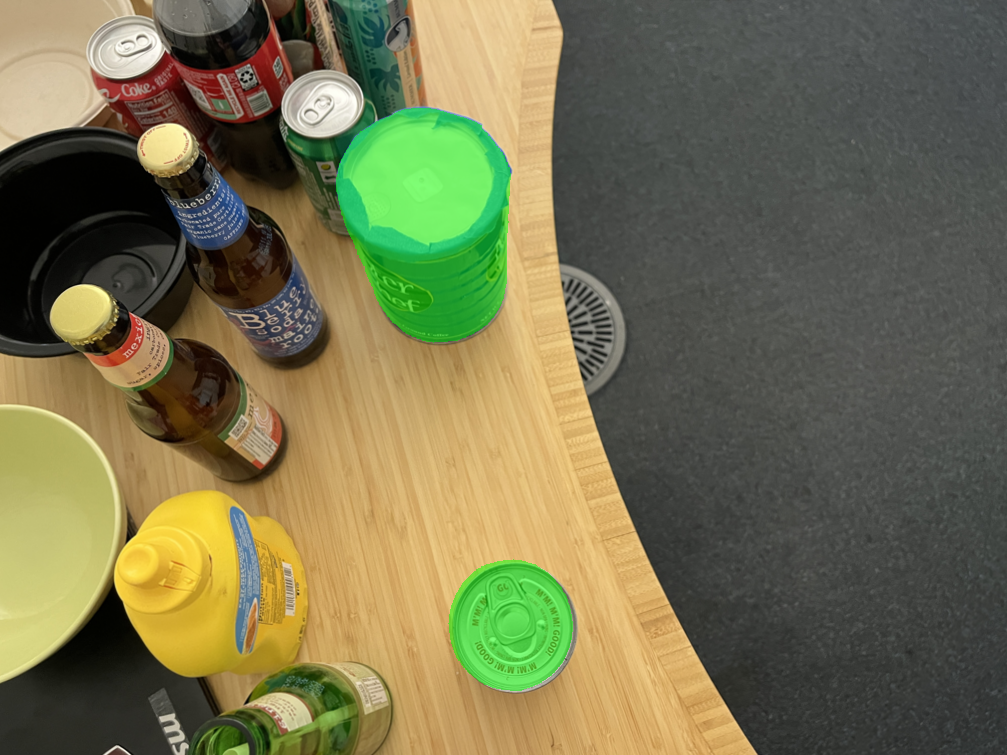}
        \caption{Swap}
    \end{subfigure}
    \begin{subfigure}[b]{0.325\linewidth}
        \includegraphics[width=0.49\linewidth]{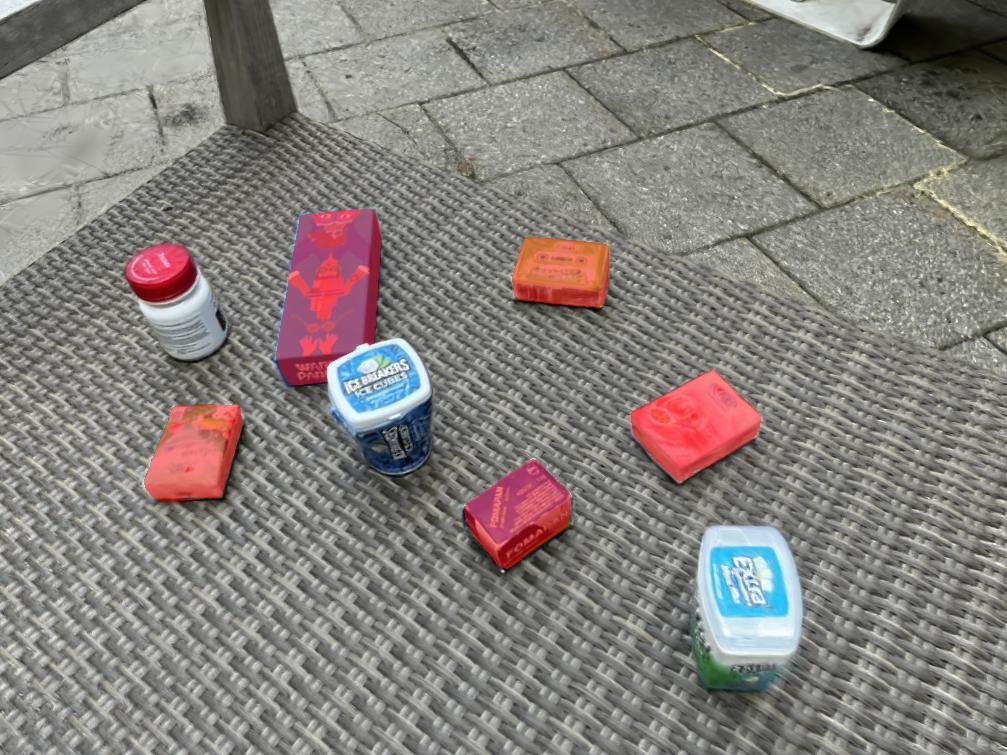}
        \includegraphics[width=0.49\linewidth]{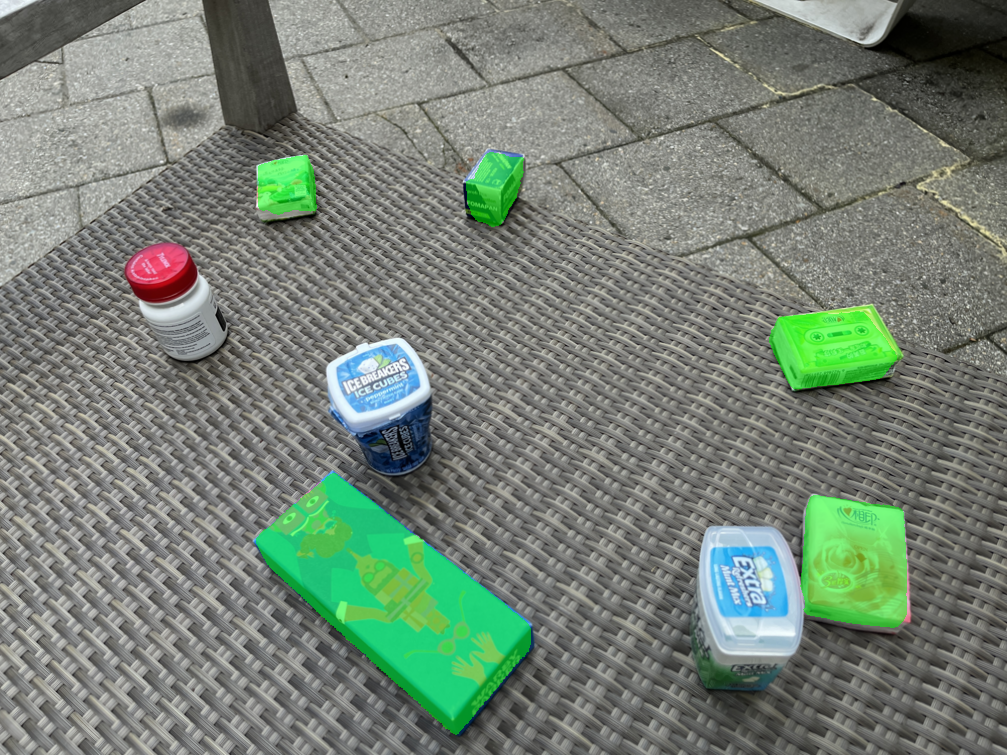}
        \caption{Bench}
    \end{subfigure}
    \begin{subfigure}[b]{0.325\linewidth}
        \includegraphics[width=0.49\linewidth]{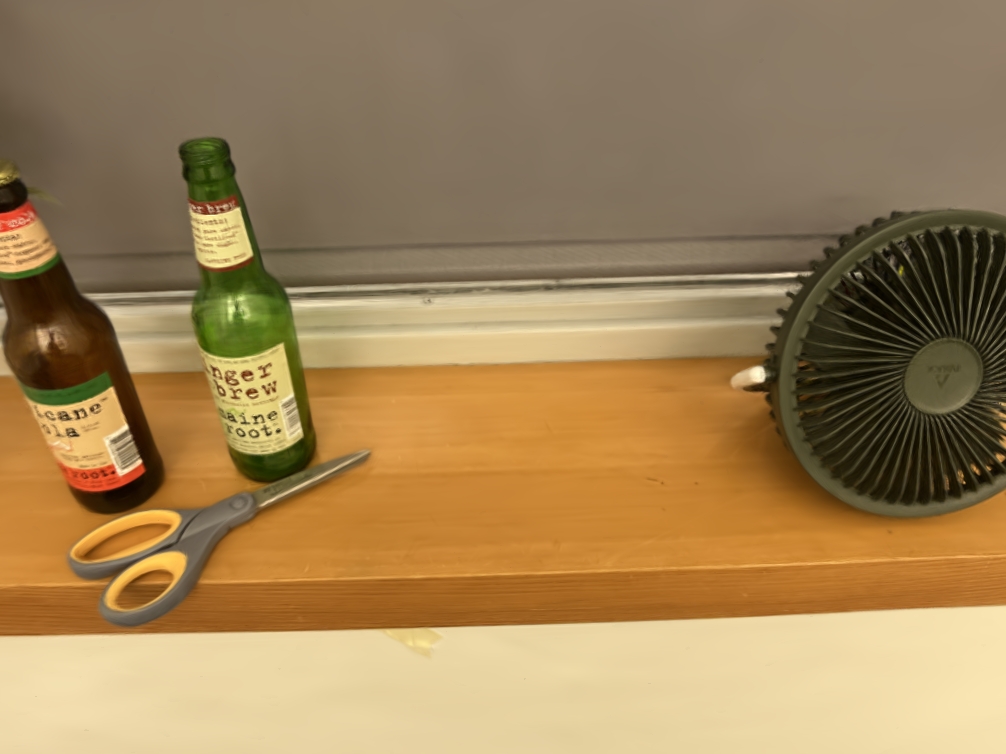}
        \includegraphics[width=0.49\linewidth]{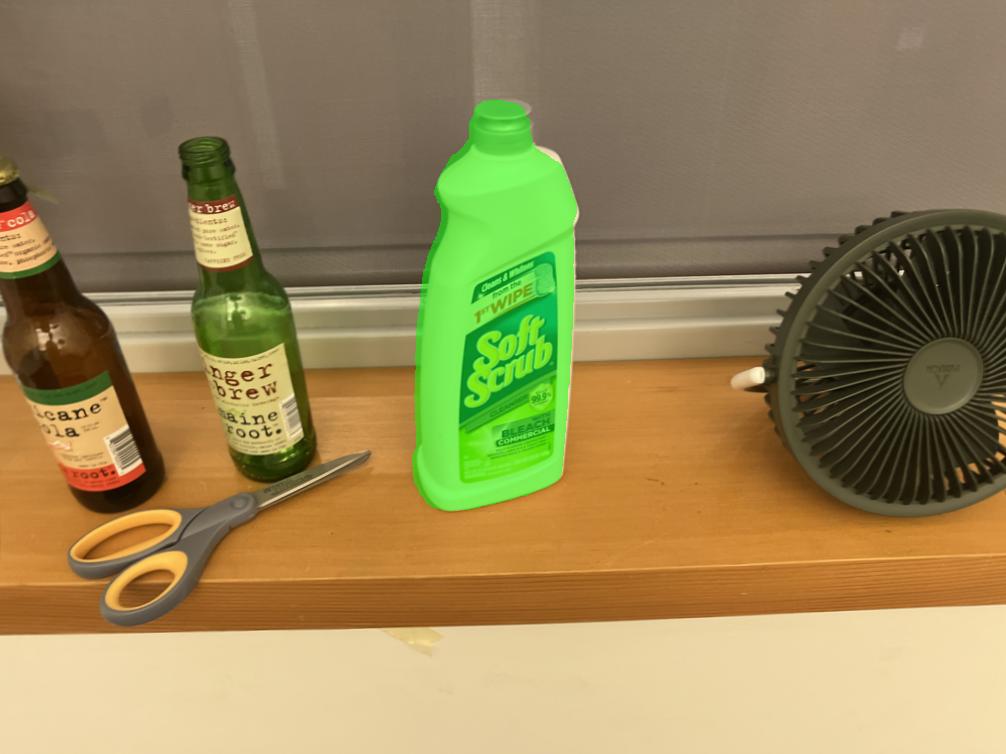}
        \caption{Sill}
    \end{subfigure}
    \begin{subfigure}[b]{0.325\linewidth}
        \includegraphics[width=0.49\linewidth]{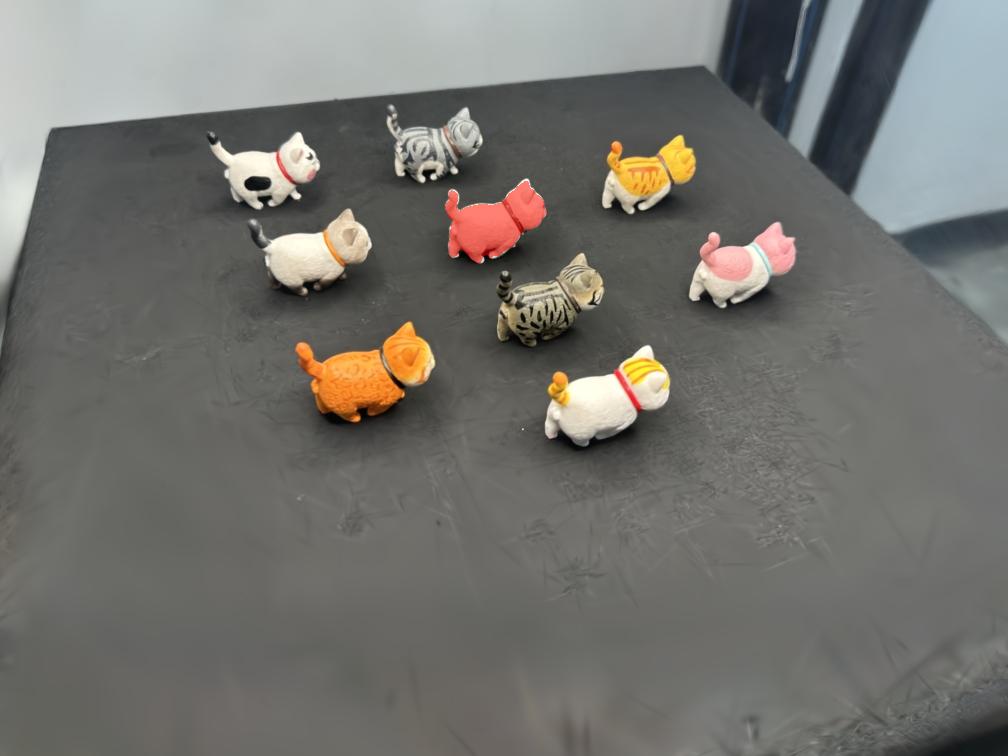}
        \includegraphics[width=0.49\linewidth]{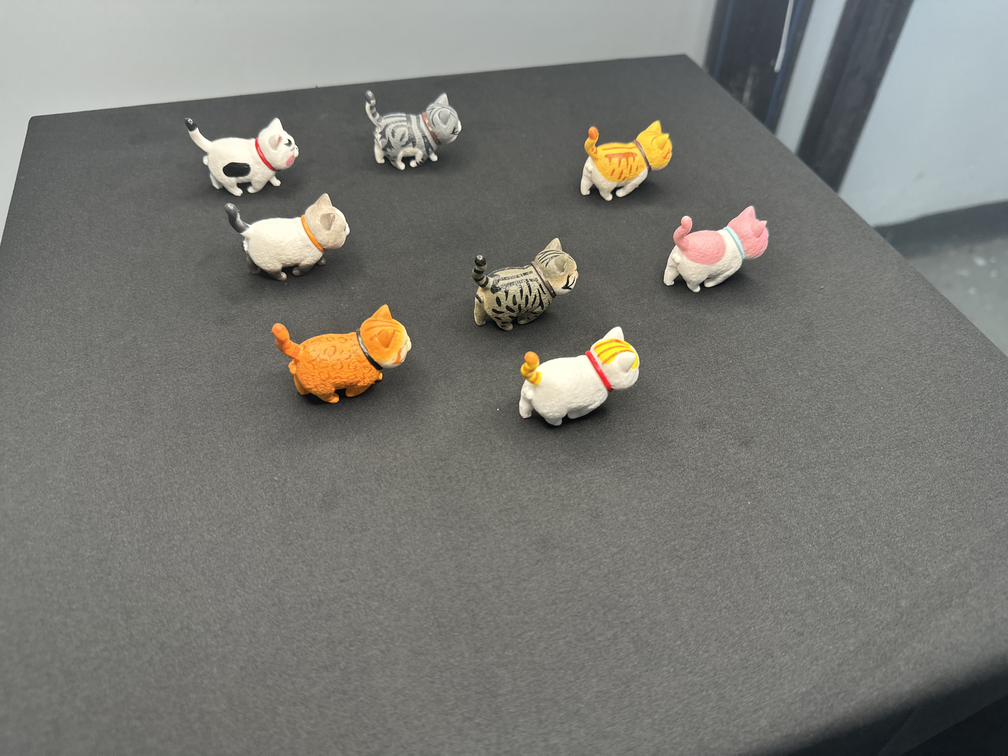}
        \caption{Cats}
    \end{subfigure}
    \caption{
        \textbf{Qualitative results}: Estimated 3D object masks projected on novel views.
        We show the projected object \textcolor{red}{move-out} (previous location) and \textcolor{green}{move-in} (new location) masks on the pre-change 3DGS render (left) and the post-change capture (right) from the same novel viewpoint for each scene.
        Please check out the supplementary video for mask projections on more views.
    }
    \label{fig:qual}
\end{figure*}

\begin{figure}[htb!]
    \begin{subfigure}[h]{0.24\linewidth}
    \includegraphics[width=\linewidth]{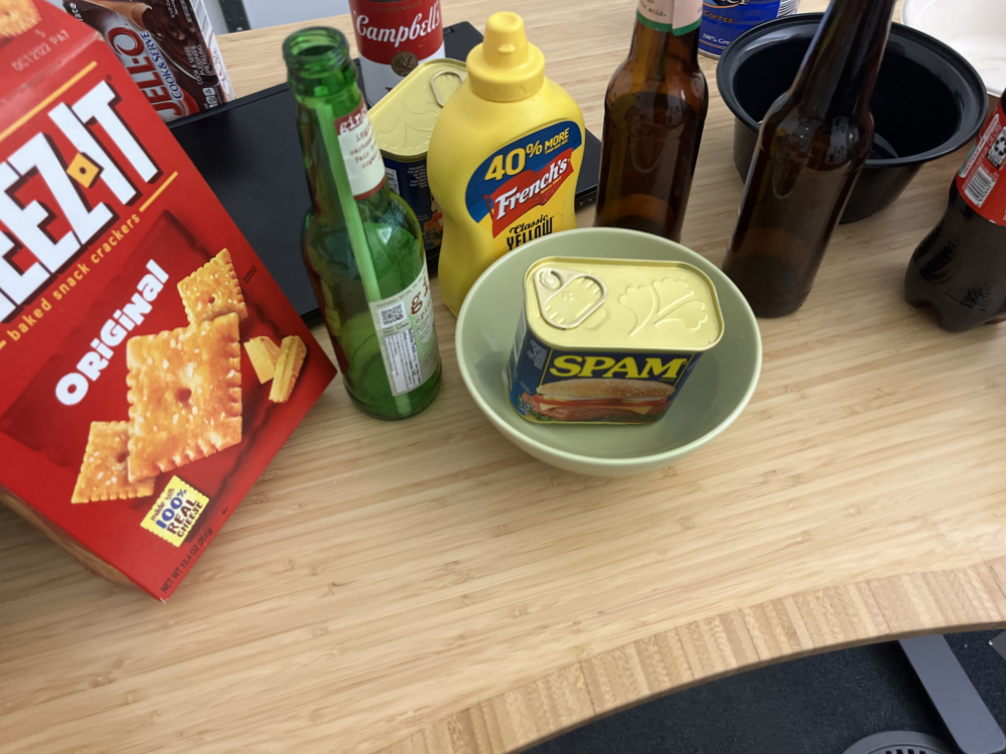}
    \caption{Eval image}
    \end{subfigure}
    \begin{subfigure}[h]{0.24\linewidth}
    \includegraphics[width=\linewidth]{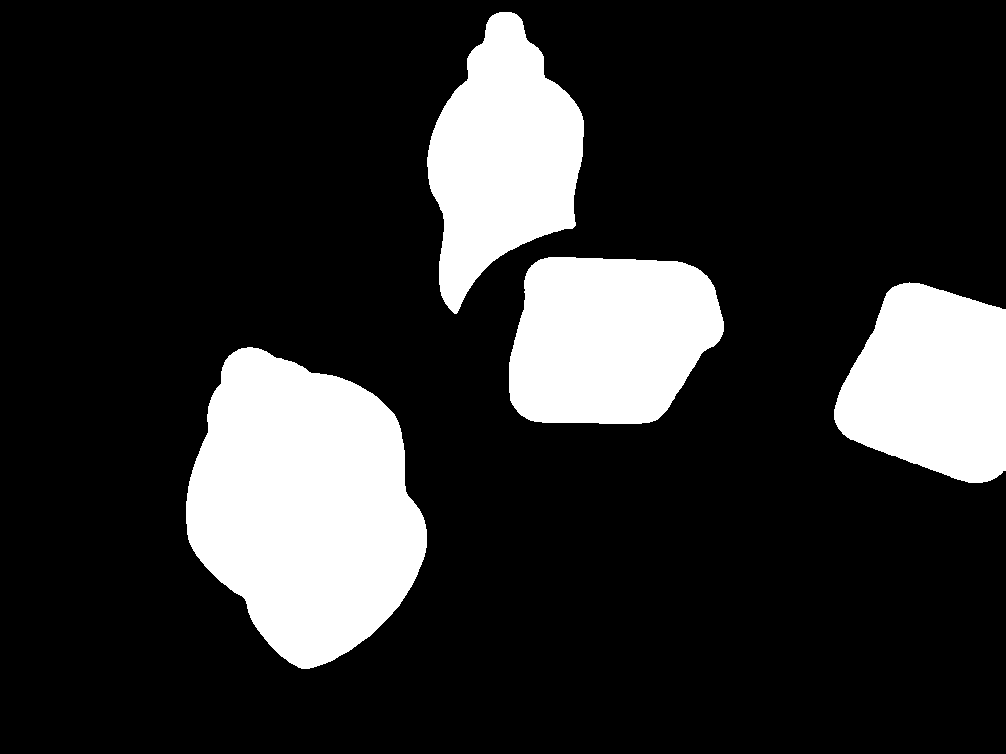}
    \caption{Ground truth}
    \end{subfigure}
    \begin{subfigure}[h]{0.24\linewidth}
    \includegraphics[width=\linewidth]{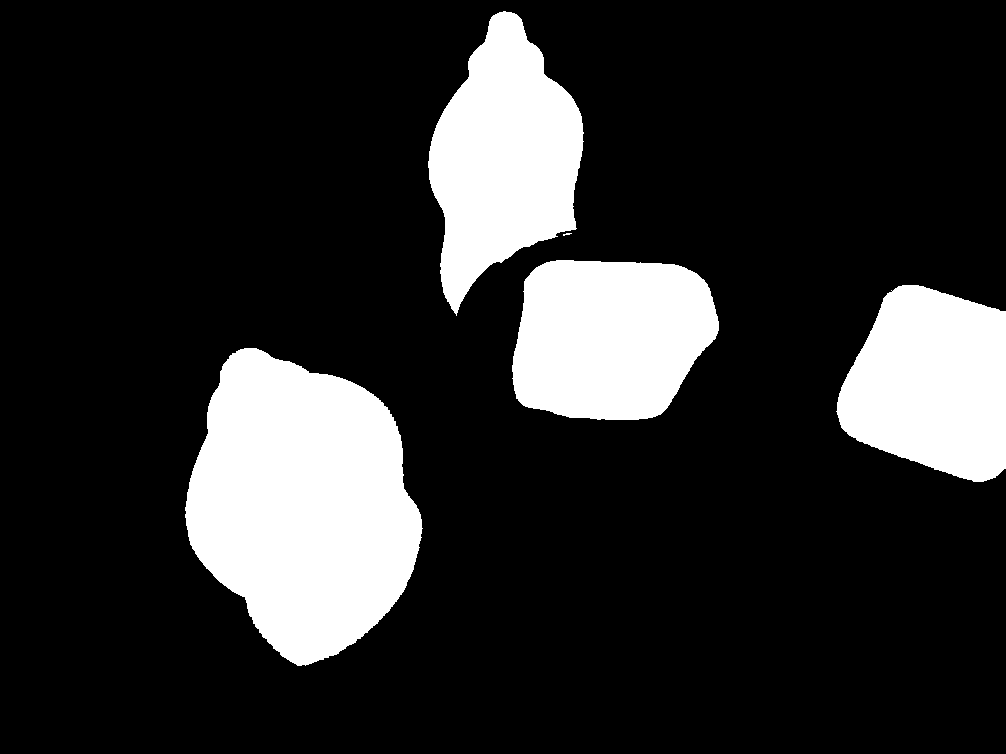}
    \caption{Ours}
    \end{subfigure}
    \begin{subfigure}[h]{0.24\linewidth}
    \includegraphics[width=\linewidth]{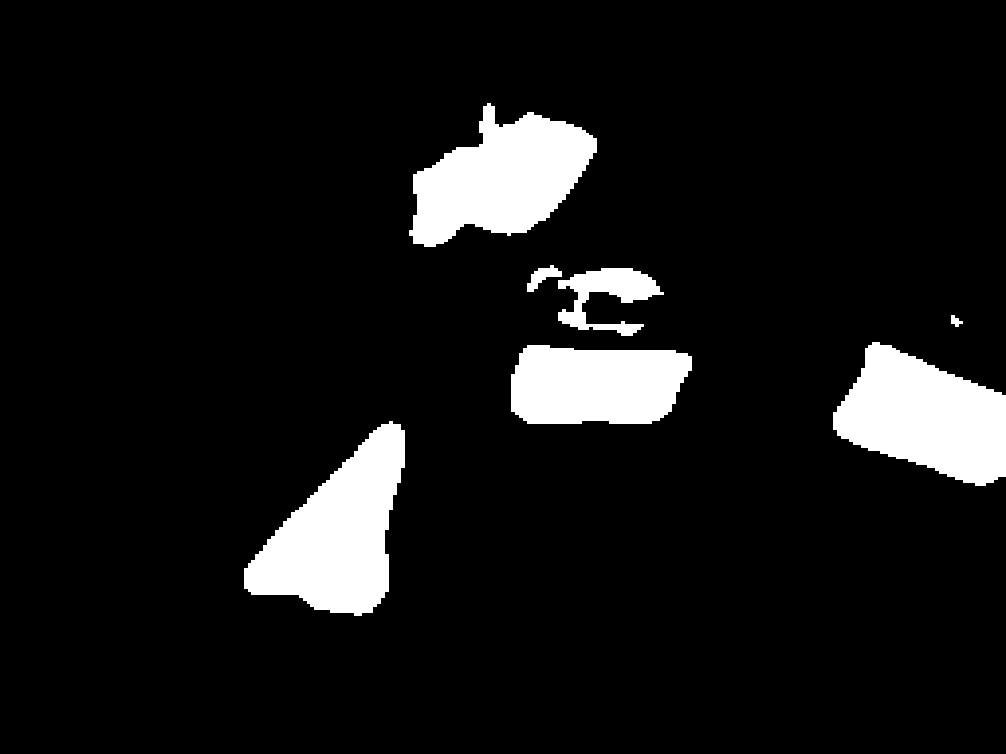}
    \caption{MTP\cite{wang2024mtp}}
    \end{subfigure}
    \caption{
        \textbf{Qualitative comparison} of our method against SOTA 2D change detection method MTP \cite{wang2024mtp} on \textit{Desk}.
    }
    \label{fig:qual2}
\end{figure}

\subsection{Global pose refinement}\label{sec:global}
We refine the initial camera and object pose change estimates with an analysis-by-synthesis approach.

With the initial estimates, we transform the 3D Gaussians in the pre-change 3DGS model,
moving those within object 3D masks according to the estimated pose changes.
We use the transformed 3D Gaussians, denoted $\mathcal{G}^{\mathcal{T}}$, to render images at the post-change views,
which are compared to the captured post-change images.
We freeze the pre-change Gaussian parameters and minimize the photometric errors between the rendered and captured images to refine the initial pose estimates:
\begin{align}
    \mathcal{T}^{*}_{c_n^\prime}, \mathcal{T}^*_k=
    \argmin_{\mathcal{T}_{c_n^\prime}, \mathcal{T}_k}
    \sum_{n=1}^N\mathcal{L}\left\{\mathcal{R}(\mathcal{T}_{c^\prime_n}; \mathcal{G}^{\mathcal{T}})\bar{\mathrm{M}}_n, I^\prime_n\bar{\mathrm{M}}_n \right\}
\end{align}
Here, $\bar{\mathrm{M}}_n$ is the inverted object move-out masks on view $n$,
used to exclude the influence of previously un-observed regions in pre-change 3DGS revealed by the object rearrangement.
$\mathcal{L}$ represents the 3DGS training loss \cite{kerbl20233d}.
The above optimization is solved using standard gradient descent.

\subsection{Occlusion-aware mask projection}\label{sec:proj}
Given a pre-change or post-change \textit{evaluation} image, 
how do we project the 3D object segmentations to its camera view in an accurate and \textit{occlusion-aware} manner?

We first estimate and refine the evaluation-view camera poses as outlined in Sec.~\ref{sec:loc} and Sec.~\ref{sec:global}.
We transform the occupied object voxels if the evaluation image is a post-change image
and project the voxels to the evaluation images to get initial masks.
Depths are then rendered within the initial masks using either the pre-change 3DGS $\mathcal{G}$ (for pre-change images) or the transformed 3DGS $\mathcal{G}^\mathcal{T}$ (for post-change images), and back-projected to 3D.
Occlusion is checked pixel-wise by determining if the back-projected depths lie inside the 3D segmentations.
The occluded pixels are reset to 0 to obtain the final evaluation-view masks.

\subsection{Implementation details}\label{sec:impl}
Our method builds on top of NeRFStudio's implementation of 3DGS: gsplat \cite{ye2024gsplat}.
We use HLoc \cite{sarlin2019coarse} for SfM (Sec.~\ref{sec:3dgs}) and camera localization (Sec.~\ref{sec:loc}).
SuperPoint+LightGlue \cite{detone2018superpoint,lindenberger2023lightglue} are used for feature detection and matching (Sec.~\ref{sec:ass}).

We conduct all our experiments on a single NVIDIA RTX 3090 GPU.

\section{Experiments}

\subsection{Datasets}
To our knowledge, the only public dataset tailored for radiance-field-based change detection is the NeRF-CD dataset \cite{huang2023c},
which focuses on simple types of changes, such as removal and insertion of objects or object components. 

To evaluate our method in more complex scenarios,
we collect a challenging real-world dataset, 3DGS-CD, 
where objects experience various types of re-arrangements including object moving, swapping and insertion in cluttered environments.
The dataset contains 5 scenes: \textit{Mustard}, \textit{Desk}, \textit{Swap}, \textit{Bench}, and \textit{Sill}.
Each scene contains about $150$ to $200$ pre-change images and $8$ post-change images,
except for \textit{Sill}, which has $60$ post-change images to aid segmentation of the inserted object.
Half of the post-change images are used for change detection, and the other half for evaluation.

For a fair comparison against the C-NeRF \cite{huang2023c} baseline,
we also evaluate our method on the first test scene \textit{Cats} of the NeRF-CD \cite{huang2023c} dataset,
which focuses on object-level changes (object removal).
Instead of using all post-change images from the scene,
we sampled 4 for change detection.
The original pre-change evaluation images provided in the dataset are used for evaluation.

\subsection{Metrics}
We evaluate our method on 2D change masks (move-in + move-out) at evaluation views,
comparing them against ground truth using Precision(\%), Recall(\%), F1(\%), and IoU(\%),
as is standard in 2D change detection.

We also report runtime performance,
which includes the time for change detection and mask projection or prediction,
but excludes NeRF or 3DGS training times.

\subsection{Baselines}
Our baseline methods include:

\textbf{MTP} \cite{wang2024mtp}: A SOTA 2D change detection method.
It is a foundation model pre-trained on remote sensing datasets for multiple tasks and fine-tuned for 2D change detection.

\textbf{C-NeRF} \cite{huang2023c}: 
A SOTA NeRF-based 3D change detection method.
C-NeRF trains two NeRFs on the pre- and post-change images in the same coordinate frame, established via joint SfM.
It identifies direction-consistent radiance differences between the two NeRFs in 3D,
and renders 2D change masks at novel views.

\textbf{C-NeRF-Diff} \cite{huang2023c}:
A baseline implemented by C-NeRF authors to compute the photometric differences between the images rendered by the pre- and post-change NeRFs. 

\textbf{D-NeRF-Diff} \cite{pumarola2021d}:
Another baseline by C-NeRF authors that trains a D-NeRF with the two image sets as two time steps.
It computes change masks using the absolute color differences between evaluation-view D-NeRF renders.

\textbf{4DGS-Diff} \cite{wu20244d}:
Similar to D-NeRF-Diff, we train 4DGS at two time steps and compute color differences between eval-view renders at different times to get the change masks.

On the 3DGS-CD dataset,
we use MTP as baseline to predict change masks on rendered and captured images at evaluation views
\footnote{C-NeRF is not used as a baseline on 3DGS-CD since it fails to train a post-change NeRF with reasonable quality using the sparse new images.}.
We adopt the \textit{ViT-B+RVSA} model fine-tuned on the CDD dataset \cite{lebedev2018change} as it performs best on our data. 
On the NeRF-CD dataset, we compare our method with C-NeRF, C-NeRF-Diff, D-NeRF-Diff and 4DGS-Diff.

\subsection{Results}

For the 3DGS-CD dataset, we present our quantitative results in Tab.~\ref{tab:exp} and qualitative results in Fig.~\ref{fig:qual} and Fig.~\ref{fig:qual2}.
Our method demonstrates high accuracy across all test cases, with all metrics consistently exceeding 90\%.
It only takes around 18s to 58s to detect 3D changes in a complex scene.

Our method is more accurate than the MTP baseline by a large margin,
despite having a slower runtime due to the inference efficiency of MTP's neural network.
While MTP's prediction accuracy could be improved through fine-tuning on in-domain data,
our method exhibits strong zero-shot performance without any fine-tuning.

\begin{table}[htb!]
\caption{
    \textbf{Quantitative evaluation} 
    of our method on 3DGS-CD dataset.
}
\label{tab:exp}
\centering
\begin{tabular}{c|c|ccccc}
\hline
Data & Methods & Precision$\uparrow$ & Recall$\uparrow$ & F1$\uparrow$ & IoU$\uparrow$ & Time\\
\hline
\multirow{2}{*}{Mustard} & \textbf{Ours} & \textbf{97.68} & \textbf{98.75} & \textbf{98.21} & \textbf{96.48} & 19s \\
 & MTP  & 94.88 & 23.08 & 37.13 & 22.80 & \textbf{1}s \\
\hline
\multirow{2}{*}{Desk} & \textbf{Ours} & \textbf{99.17} & \textbf{95.82} & \textbf{97.47} & \textbf{95.06} & 21s  \\
 & MTP  & 95.69 & 34.38 & 50.58 & 33.85 & \textbf{1}s \\
\hline
\multirow{2}{*}{Swap} & \textbf{Ours} & \textbf{98.84} & \textbf{98.82} & \textbf{98.83} & \textbf{97.69} & 28s \\
 & MTP  & 94.19 & 24.61 & 39.03 & 24.25 & \textbf{1}s \\
\hline
\multirow{2}{*}{Bench} & \textbf{Ours} &\textbf{96.51} &\textbf{96.76} & \textbf{96.64} & \textbf{93.49} & 58s \\
 & MTP  & 90.19 & 88.74 & 89.46 & 80.93 & \textbf{1}s \\
\hline
\multirow{2}{*}{Sill} & \textbf{Ours} & \textbf{95.00} & \textbf{97.86} & \textbf{96.41} &  \textbf{93.07} & 18s\\
 & MTP  & 48.25 & 30.84 & 37.63 & 23.17 & \textbf{2}s \\
\hline
\end{tabular}
\end{table}

We further report our results on the ``Cats'' scene from the NeRF-CD dataset in Tab.~\ref{tab:nerfcd} and Fig.~\ref{fig:qual}(f).
Our method is up to 14\% more accurate and 556 times faster than C-NeRF.
This significant speedup is primarily due to C-NeRF's reliance on computationally expensive ray tracing to identify changes and render masks,
whereas our method leverages EfficientSAM, real-time 3DGS rendering and fast classic computer vision techniques to build our pipeline.

Our method also outperforms the dynamic NeRF/3DGS-based baselines (i.e. D-NeRF-Diff, 4DGS-Diff) by a large margin in change detection accuracy.
This is expected since the dynamic NeRF/3DGS models are not designed or optimized for predicting accurate 3D changes.
We do not include a runtime performance comparison with these baselines because they don't have an explicit change detection step.
Comparing our method’s runtime to, for instance, the training time of 4DGS, would not be a fair comparison.

\begin{table}[htb!]
\caption{
    \textbf{Quantitative evaluation} 
    of our method on the test scene \textit{Cats} from the NeRF-CD \cite{huang2023c} dataset.
    Runtimes for the *-Diff baselines are omitted as they don't have an explicit change detection step.
}
\label{tab:nerfcd}
\centering
\begin{tabular}{c|ccccc}
\hline
Methods & Precision$\uparrow$ & Recall$\uparrow$ & F1$\uparrow$ & IoU$\uparrow$ & Time \\
\hline
\textbf{3DGS-CD} & \textbf{97.94} & \textbf{95.62} & \textbf{96.76} & \textbf{93.73} & \textbf{22s} \\
C-NeRF & 86.70 & 90.46 & 88.54 & 79.47 & 3.4\textbf{h}\\
C-NeRF-Diff & 58.80 &  83.50 &  68.86 &  52.70 & - \\
D-NeRF-Diff & 59.24 & 60.57  & 59.68  & 42.71 & - \\
4DGS-Diff & 62.44 & 90.73 & 73.97 & 58.69 & - \\
\hline
\end{tabular}
\end{table}

\subsection{Ablation study}\label{sec:ablate}
\subsubsection{Number of post-change images}
\begin{figure}[htb!]
    \centering
    \includegraphics[width=\linewidth]{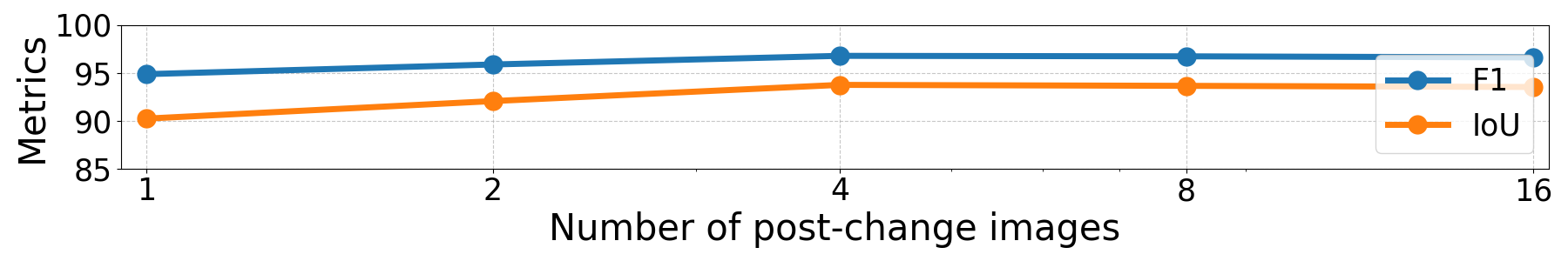}
    \caption{
        \textbf{Ablation study} on the number of post-change images on \textit{Cats}.
        Our method maintains consistent performance across different numbers of post-change images.
    }
    \label{fig:ablate_num}
\end{figure}

How does our method scale to fewer or more post-change images?
We sample and use different numbers of post-change images from the \textit{Cats} scene for change detection,
and evaluate our method with the same evaluation images.
As Fig.~\ref{fig:ablate_num} shows, our method maintains stable performance across varying numbers of post-change images,
even when using just one image.
The slight decline in accuracy with fewer images is mainly due to the incomplete object template built from limited views,
which degrades the quality of the box prompt for EfficientSAM on pre-change images (Sec.~\ref{sec:seg}).

\section{Applications}

Our method significantly advances the accuracy and efficiency of radiance-field-based 3D change detection,
enabling a wide range of real-world applications.
Below, we show three examples to demonstrate its potential use cases.

\subsection{Object removal as prompt for object reconstruction}\label{sec:recon}

The emergence of NeRF and 3DGS has made the task of 3D object reconstruction more accessible to non-professional users.
With just a phone, people can now casually capture RGB images to build an object-only NeRF or 3DGS model in minutes.
For complex objects in cluttered environments, however,
user prompts are often needed to consistently isolate the same foreground object on the input images.

Common prompts like clicks and language can be ambiguous in challenging scenarios.
For instance, click prompts, particularly single clicks, struggle with multi-level granularity
where a click may represent an object component or the entire object.
Language prompts can also lead to confusion, especially when multiple similar objects are present.

\begin{figure}[htb!]
    \centering
    \begin{subfigure}[h]{0.18\linewidth}
        \centering
        \begin{tikzpicture}
            \node[anchor=south west, inner sep=0] (image) at (0,0) {
                \includegraphics[trim=380 90 350 90, clip, width=\linewidth]{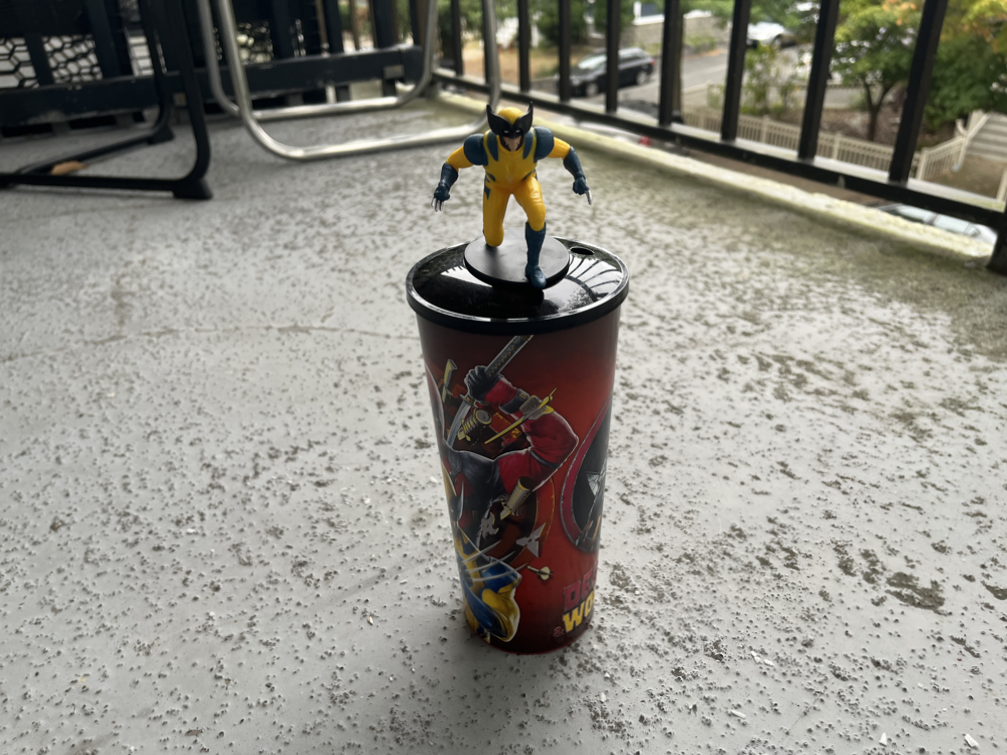}
            };
            \begin{scope}[x={(image.south east)}, y={(image.north west)}]
                \fill[red] (0.48, 0.93) circle (2pt); 
                \fill[blue] (0.5, 0.5) circle (2pt); 
                \fill[green] (0.5, 0.4) circle (2pt); 
                \fill[green] (0.48, 0.85) circle (2pt); 
                \fill[green] (0.48, 0.71) circle (2pt); 
                \fill[green] (0.3, 0.65) circle (2pt); 
            \end{scope}
        \end{tikzpicture}
        \caption{Clicks}
    \end{subfigure}
    \begin{subfigure}[h]{0.18\linewidth}
        \centering
        \includegraphics[trim=380 90 345 90, clip, width=\linewidth]{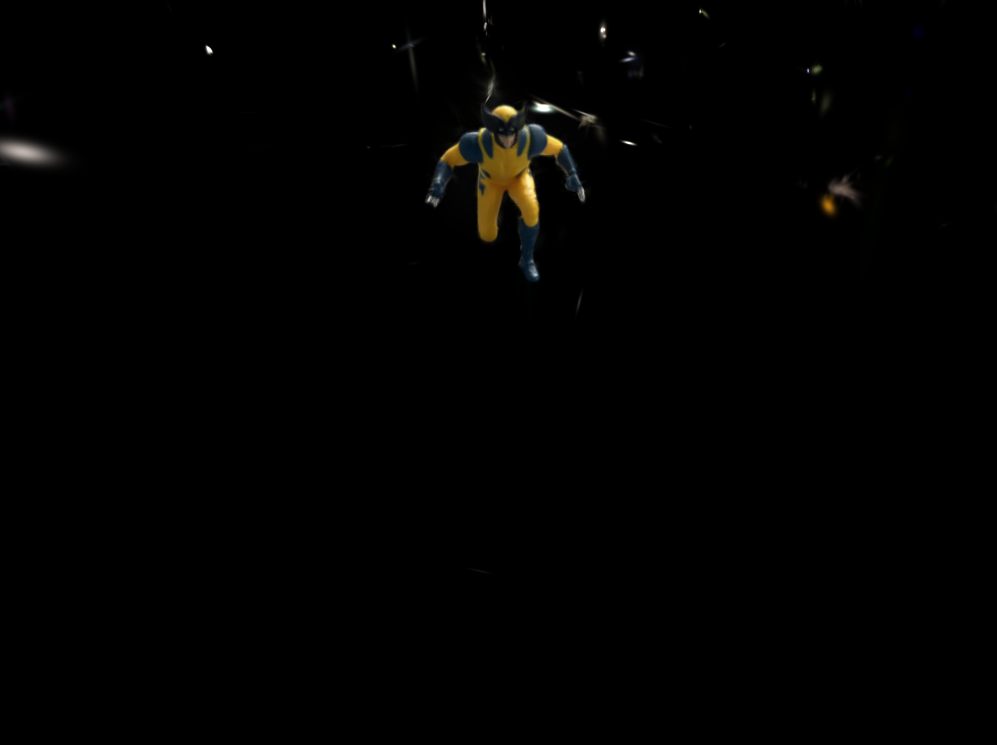}
        \caption{{\color{red}\scalebox{1.5}\textbullet}}
    \end{subfigure}
    \begin{subfigure}[h]{0.18\linewidth}
        \centering
        \includegraphics[trim=380 90 345 90, clip, width=\linewidth]{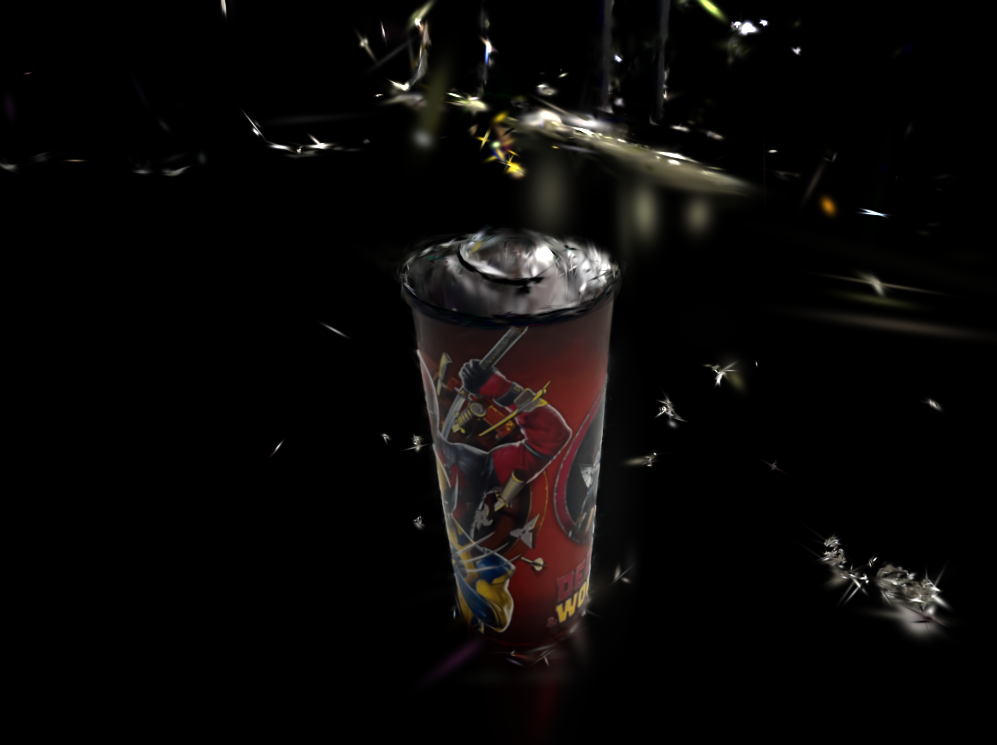}
        \caption{{\color{blue}\scalebox{1.5}\textbullet}}
    \end{subfigure}
    \begin{subfigure}[h]{0.18\linewidth}
        \centering
        \includegraphics[trim=380 90 345 90, clip, width=\linewidth]{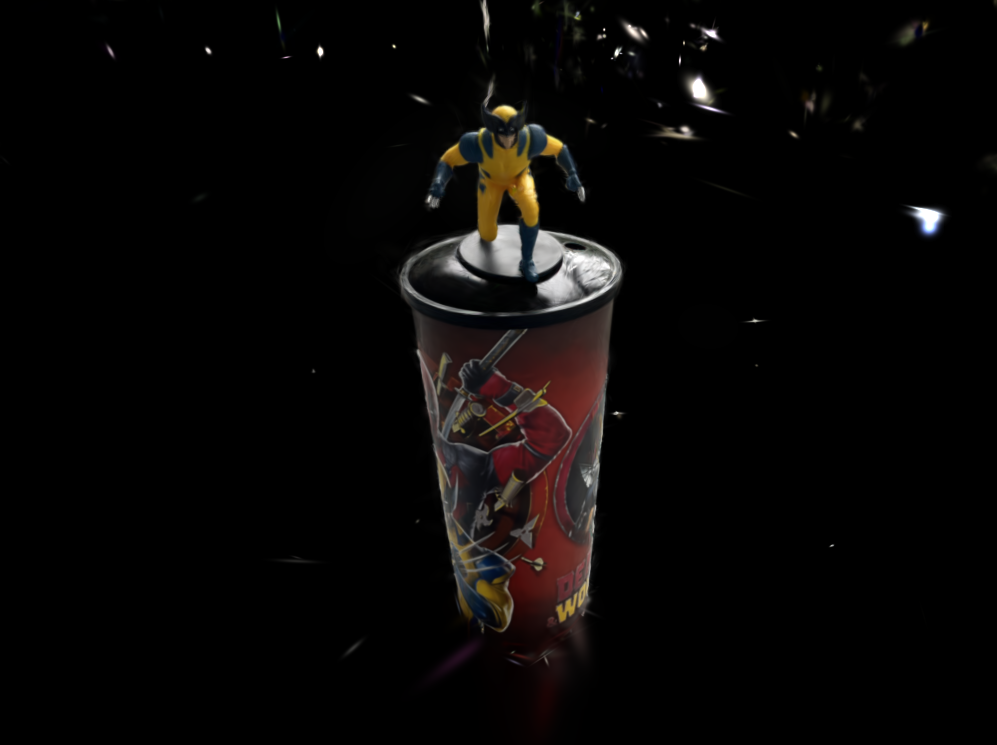}
        \caption{{\color{green}\scalebox{1.5}\textbullet}}
    \end{subfigure}
    \begin{subfigure}[h]{0.18\linewidth}
        \centering
        \includegraphics[trim=380 90 345 90, clip, width=\linewidth]{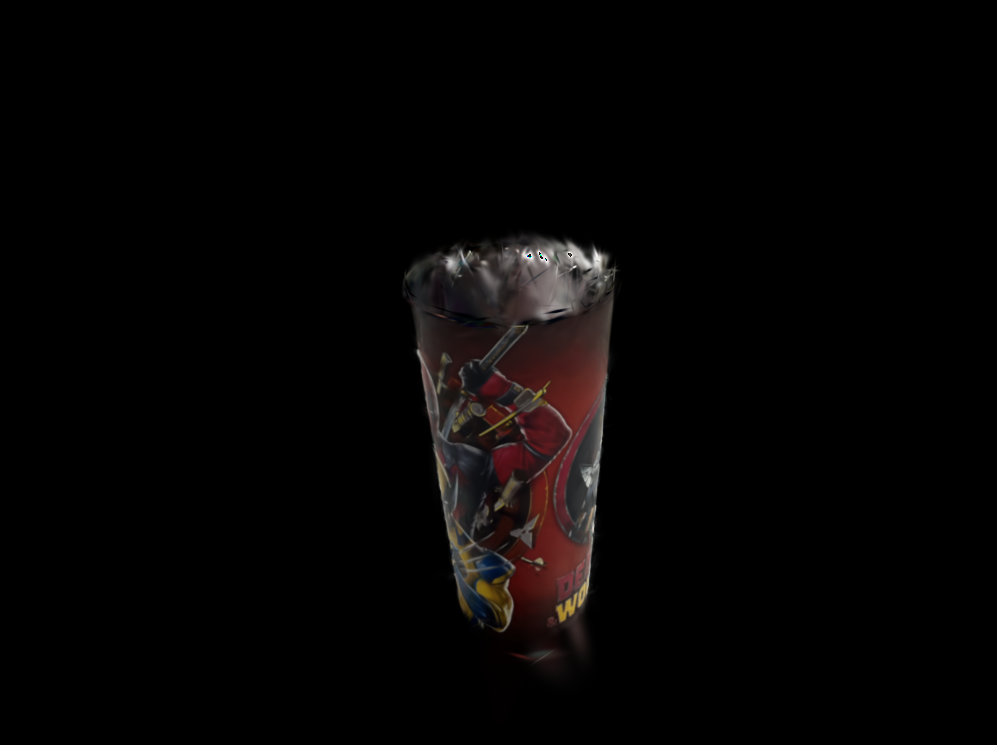}
        \caption{Language}
    \end{subfigure}
    \caption{
        Click or language prompts can be ambiguous for object extraction:
        SAGA \cite{cen2023segment} struggles to segment the entire object with single (a) click prompts on (b) the figurine or (c) the cup body.
        We had to use (d) multiple precise click prompts to extract the complete object.
        Using GPT-4o generated language prompt 
        ``\textit{Coke cup with a wolverine figurine on the lid}"
        also failed to isolate the whole object.
    }
    \label{fig:cup_saga}
\end{figure}
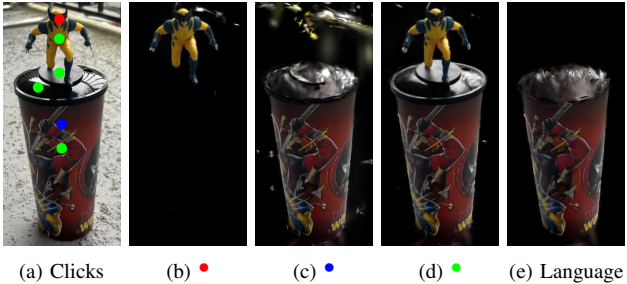

To illustrate this, we use the SOTA prompt-based 3DGS segmentation method SAGA \cite{cen2023segment} to isolate a target object from its 3DGS-represented scene,
as shown in Fig.~\ref{fig:cup_saga}.
While SAGA explicitly accounts for multi-level granularity,
it still struggles to segment the entire object with a single click prompt ({\color{red}\scalebox{1.5}\textbullet} or {\color{blue}\scalebox{1.5}\textbullet}),
even with the best thresholds.
We had to provide multiple precise clicks ({\color{green}\scalebox{1.5}\textbullet}) to extract the complete object.
However, providing such precise clicks can be difficult on mobile devices.
Similarly, using a language prompt generated by GPT-4o also failed to isolate the whole object.

\begin{figure}[htb!]
    \begin{subfigure}[h]{0.40\linewidth}
        \centering
        \includegraphics[width=\linewidth]{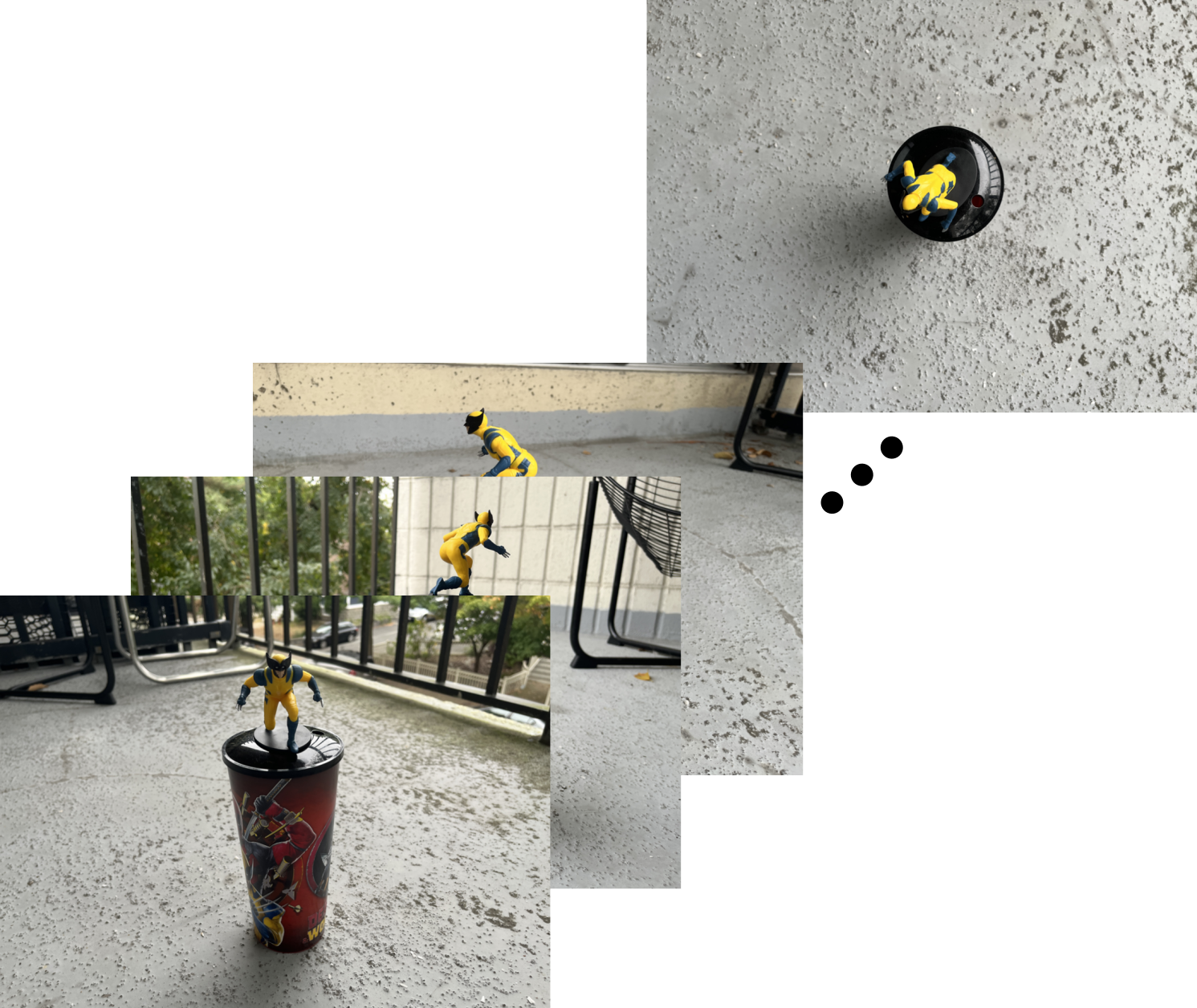}
        \caption{RGB video scan}
    \end{subfigure}
    \begin{subfigure}[h]{0.40\linewidth}
        \centering
        \includegraphics[width=\linewidth]{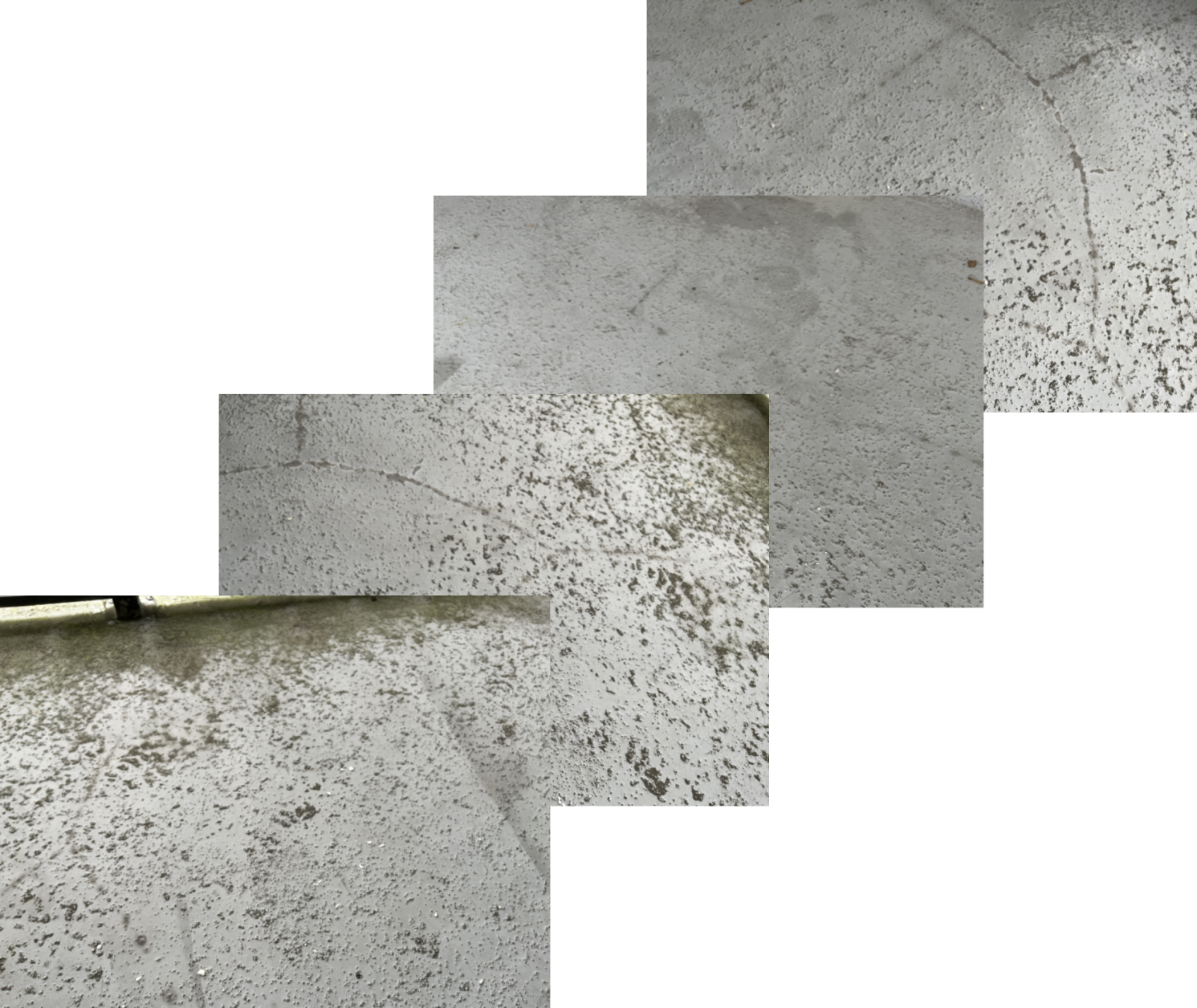}
        \caption{Object removal prompt}
    \end{subfigure}
    \begin{subfigure}[h]{0.17\linewidth}
        \centering
        \includegraphics[trim=370 90 350 100, clip, width=\linewidth]{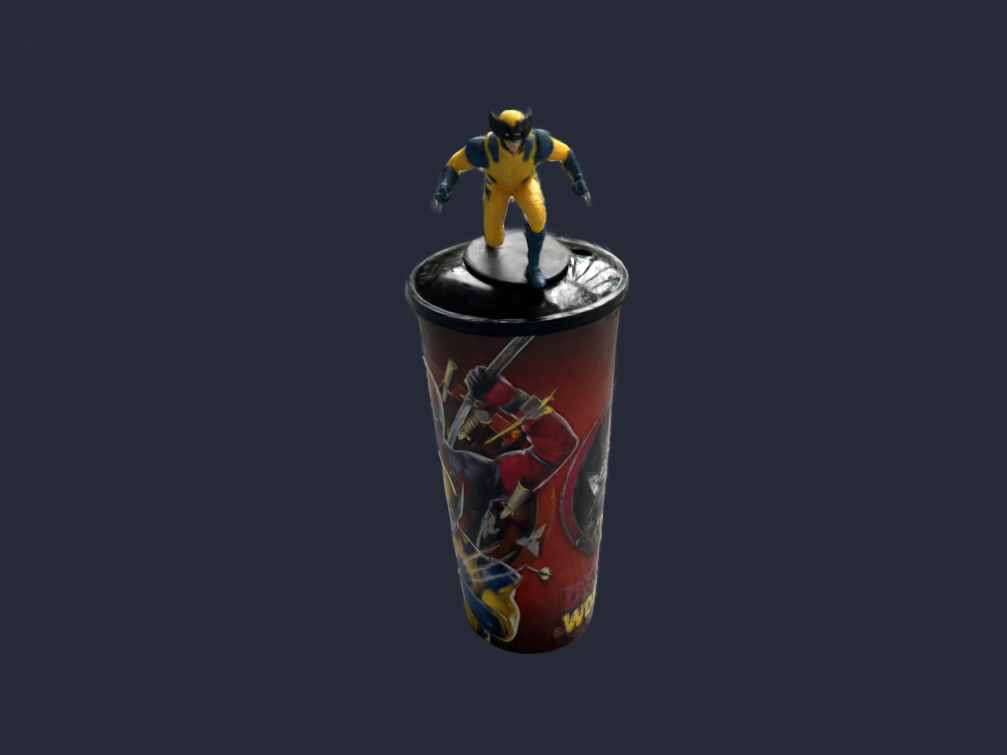}
        \caption{Obj-GS}
    \end{subfigure}
    \caption{
        \textbf{Object removal as prompt} for 3D object reconstruction:
        Our method only requires a few more images for the object-removed scene to reconstruct the target object without ambiguity.
    }
    \label{fig:cup_ours}
\end{figure}

To address this, we propose using \textbf{object removal as an unambiguous prompt},
as shown in Fig.~\ref{fig:cup_ours}.
After capturing images of the object in its scene,
users only have to remove the object, and take a few more images for the object-removed scene.
Our method can then consistently identify the target object from the pre-change images,
allowing us to train an object-3DGS without ambiguity.
Compared to the click or language prompts,
the object removal prompt is more intuitive and requires minimal user efforts.

\subsection{Robot workspace reset} \label{sec:reset}

\begin{figure}
    \centering
    \begin{subfigure}[b]{0.33\linewidth}
        \includegraphics[width=\linewidth]{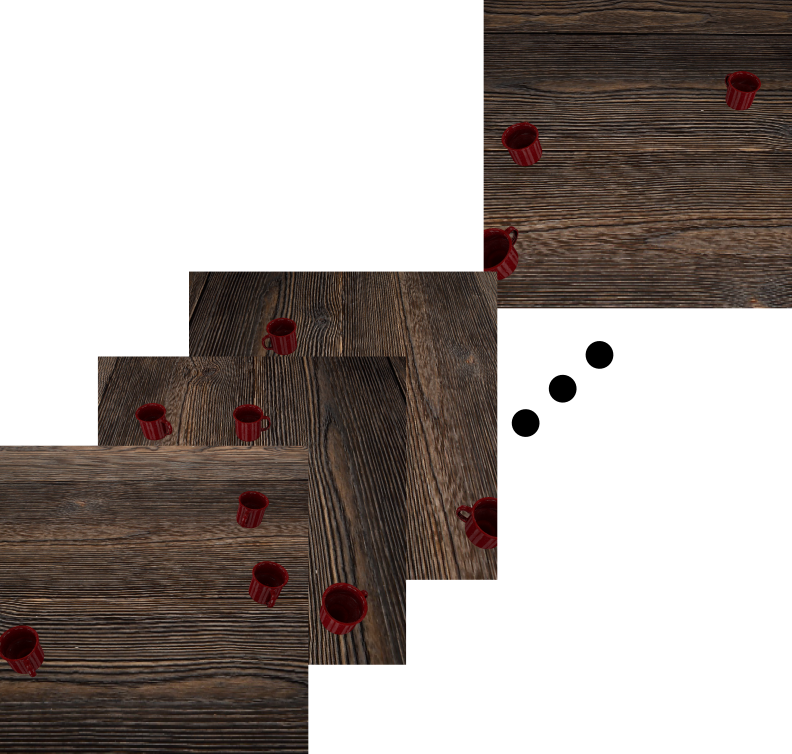}
        \caption{Pre-change images}
    \end{subfigure}
    \begin{subfigure}[b]{0.33\linewidth}
        \includegraphics[width=\linewidth]{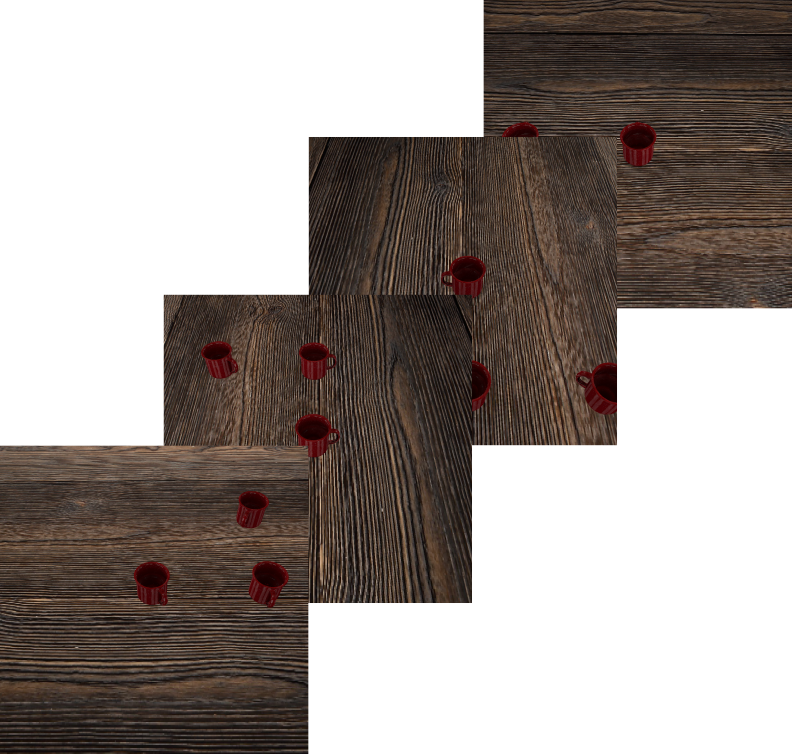}
        \caption{Post-change images}
    \end{subfigure}
    \begin{subfigure}[b]{0.30\linewidth}
        \href{
            https://drive.google.com/file/d/1Q5rPPRjIEjEGLQul_UjiomD9HYLv07ny/view?usp=drive_link
        }{\includegraphics[width=\linewidth]{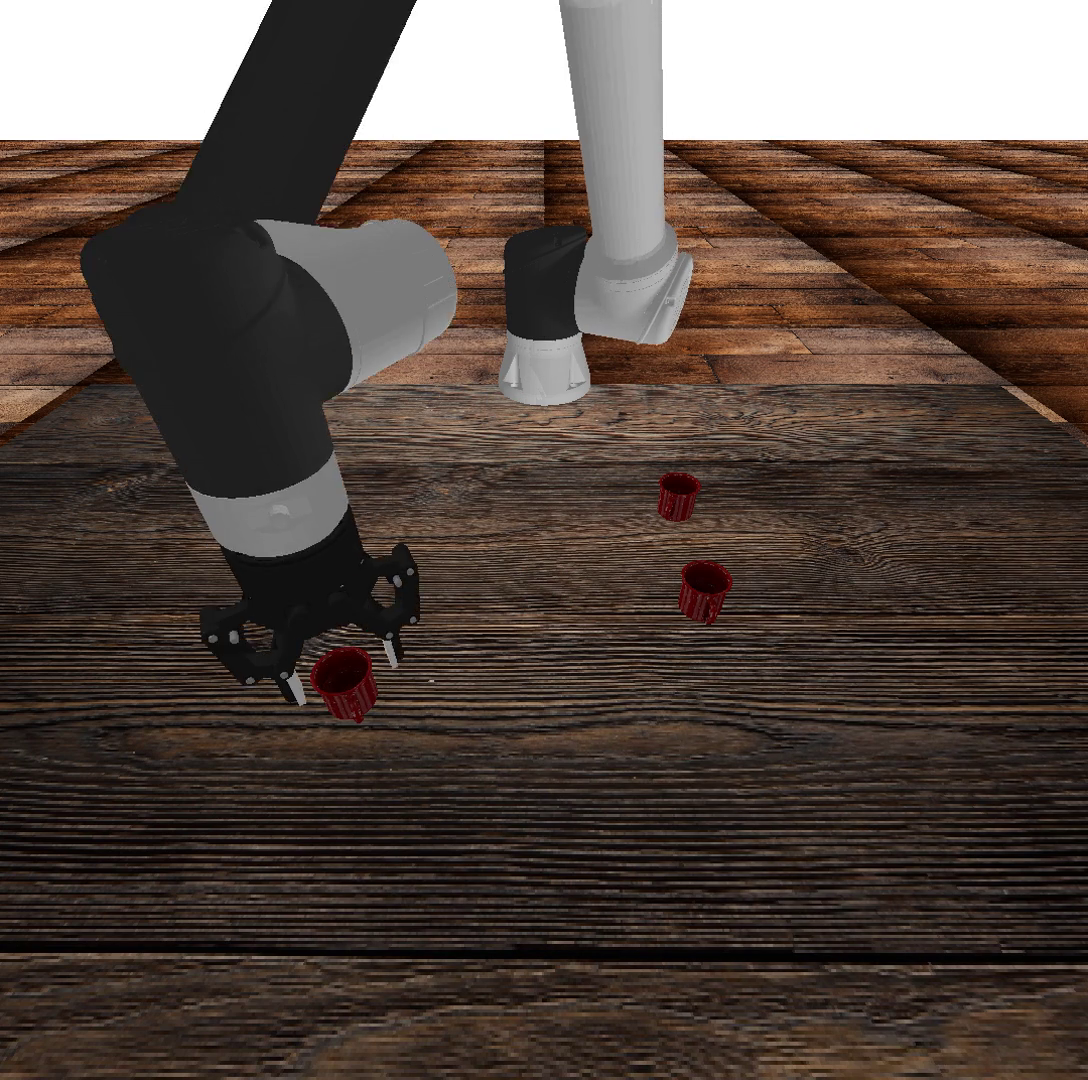}}
        \caption{Robot reset}
    \end{subfigure}
    \caption{
        3DGS-CD enables \textbf{robot workspace reset}.
        Please click on the sub-figure (c) for the robot reset video.
    }
    \label{fig:reset}
\end{figure}

Our method precisely estimates the shapes and pose changes of moved objects,
enabling a robotic manipulator to reset the scene to its initial state. 
This capability is particularly useful in applications such as warehouse or housekeeping robots,
where resetting the workspace after operation can be crucial. 

The robot workspace reset process can be summarized as follows:
(1)~RGB video scan to capture the initial state of the robot workspace;
(2)~Move objects in the workspace;
(3)~Capture a few more RGB images for the changed state of the workspace;
(4)~Use our method to estimate the moved objects' shapes and pose changes in under 30s;
(5)~Robot pick and place to reset the state of the scene.

Key advantages of our robot workspace reset method over existing solutions (e.g. SayCan~\cite{ahn2022can}) include:
(1)~No depth sensor or depth estimator is required.
(2)~No need for object class, model or object detectors.
(3)~No need for potentially ambiguous language prompts
-- Language prompts for a workspace with multiple identical objects present could confuse many existing models (e.g. SayCan \cite{ahn2022can}).
(4)~\textit{Precise} reset of the object position and orientation.

Figure~\ref{fig:reset} shows a simulated experiment we conducted in PyBullet \cite{coumans2016pybullet},
where a robot arm resets the pose of a mug on a table.
In this experiment, we used forward kinematics,
rather than structure-from-motion (SfM) or visual localization,
to estimate the pre- and post-change poses of the robot's RGB camera.
Without the SfM point cloud, the pre-change 3D Gaussians are randomly initialized before optimization.
Once the middle mug's pose change and shape are computed,
we determine the pick-and-place poses for the robot gripper and use inverse kinematics to compute the joint motions.
For simplicity, we use the initial gripper orientation as the grasp orientation and use the object's centroid as the grasp position.

Although our demo simplifies the grasp pose generation and collision avoidance process,
the object change information our method provides can potentially enable more advanced motion planning algorithms.

\subsection{Fast sparse-view-guided 3DGS update}\label{sec:update}
\begin{figure*}[htb!]
    \centering
    \begin{subfigure}[b]{0.16\linewidth}
        \includegraphics[width=\linewidth]{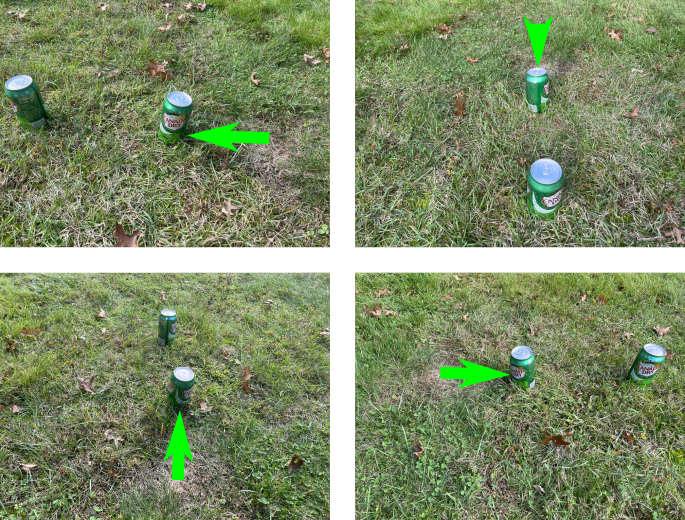}
        \caption{RGB observations}
    \end{subfigure}
    \begin{subfigure}[b]{0.16\linewidth}
        \begin{tikzpicture}
            \node[anchor=south west,inner sep=0] (image) at (0,0) {
                \includegraphics[width=\linewidth]{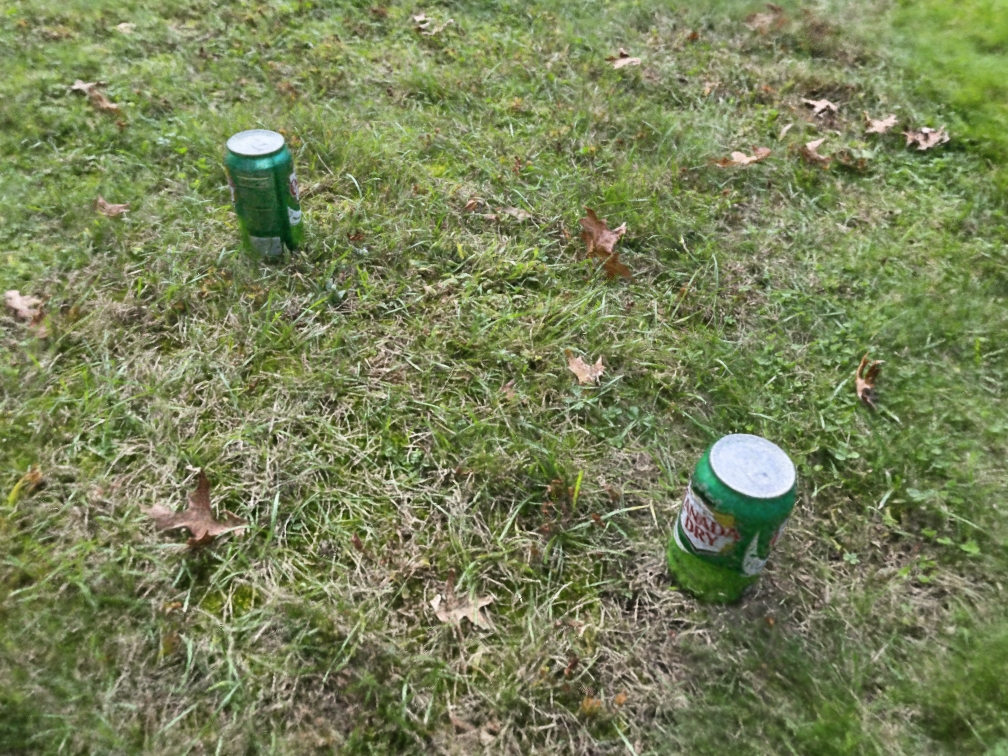}
            };
            \begin{scope}[x={(image.south east)},y={(image.north west)}]
                \draw[red,thick, dashed] (0.64, 0.45) rectangle (0.81,0.17);
                \draw[green, thick, dashed] (0.45, 0.65) rectangle (0.57, 0.43);
            \end{scope}
        \end{tikzpicture}
        \caption{Pre-trained}
    \end{subfigure}
    \begin{subfigure}[b]{0.16\linewidth}
        \begin{tikzpicture}
            \node[anchor=south west,inner sep=0] (image) at (0,0) {
                \includegraphics[width=\linewidth]{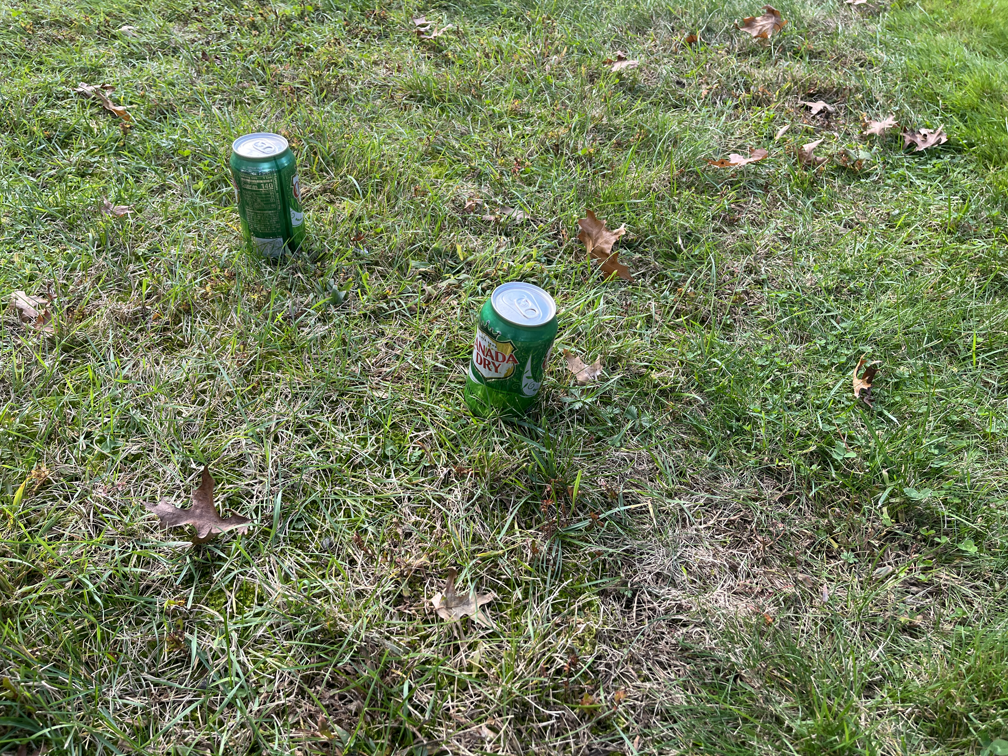}
            };
            \begin{scope}[x={(image.south east)},y={(image.north west)}]
                \draw[red,thick, dashed] (0.64, 0.45) rectangle (0.81,0.17);
                \draw[green, thick, dashed] (0.45, 0.65) rectangle (0.57, 0.43);
            \end{scope}
        \end{tikzpicture}
        \caption{Ground truth}
    \end{subfigure}
    \begin{subfigure}[b]{0.16\linewidth}
        \begin{tikzpicture}
            \node[anchor=south west,inner sep=0] (image) at (0,0) {
                \includegraphics[width=\linewidth]{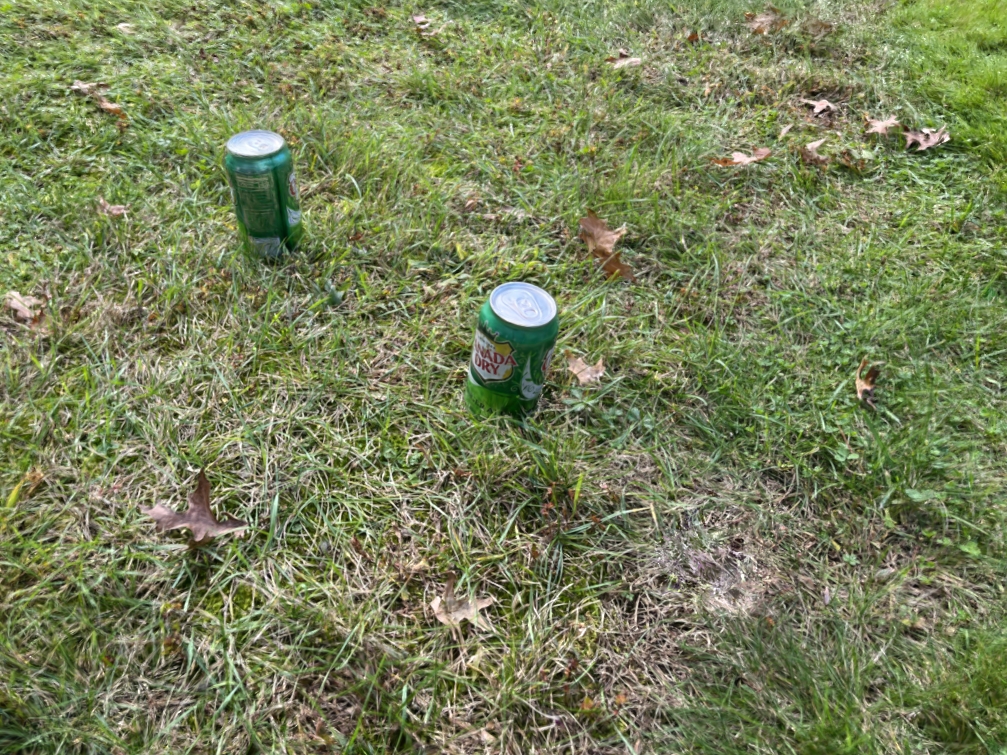}
            };
            \begin{scope}[x={(image.south east)},y={(image.north west)}]
                \draw[red,thick, dashed] (0.64, 0.45) rectangle (0.81,0.17);
                \draw[green, thick, dashed] (0.45, 0.65) rectangle (0.57, 0.43);
            \end{scope}
        \end{tikzpicture}
        \caption{3DGS-Update (Ours)}
    \end{subfigure}
    \begin{subfigure}[b]{0.16\linewidth}
        \begin{tikzpicture}
            \node[anchor=south west,inner sep=0] (image) at (0,0) {
                \includegraphics[width=\linewidth]{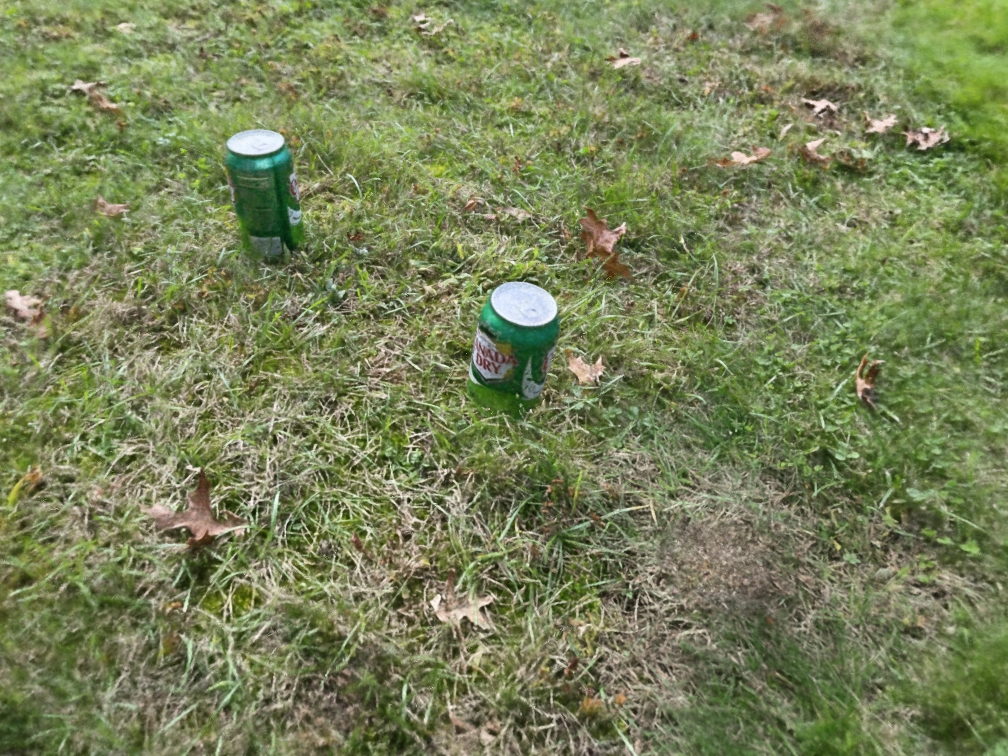}
            };
            \begin{scope}[x={(image.south east)},y={(image.north west)}]
                \draw[red,thick, dashed] (0.64, 0.45) rectangle (0.81,0.17);
                \draw[green, thick, dashed] (0.45, 0.65) rectangle (0.57, 0.43);
            \end{scope}
        \end{tikzpicture}
        \caption{NeRF-Update \cite{lu2024fast}}
    \end{subfigure}
    \begin{subfigure}[b]{0.16\linewidth}
        \begin{tikzpicture}
            \node[anchor=south west,inner sep=0] (image) at (0,0) {
                \includegraphics[width=\linewidth]{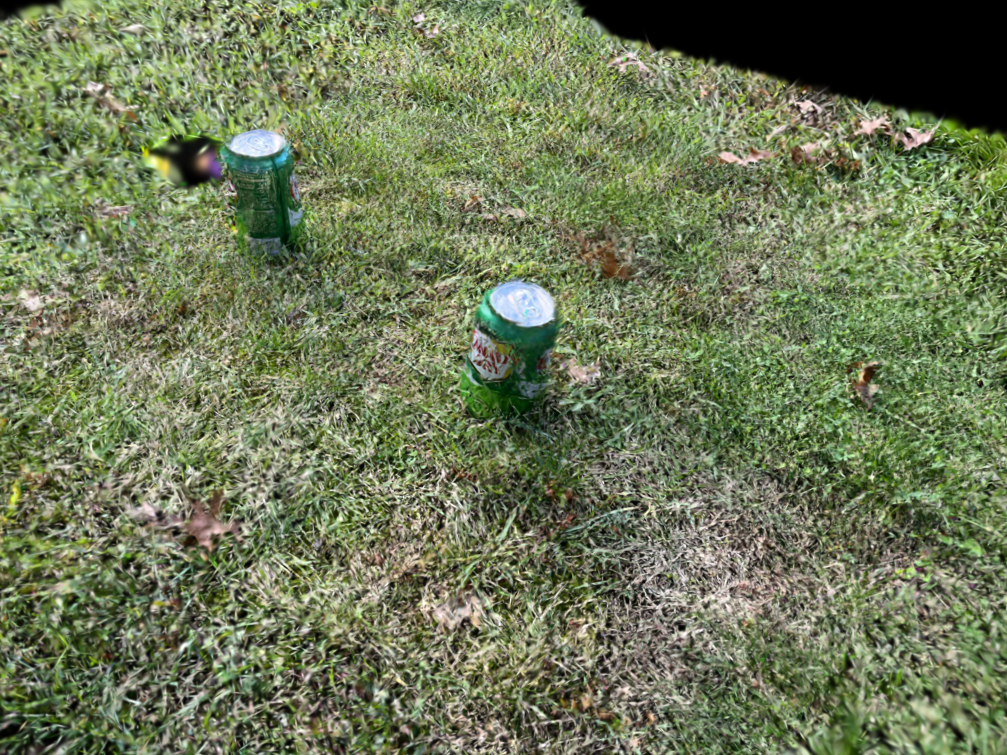}
            };
            \begin{scope}[x={(image.south east)},y={(image.north west)}]
                \draw[red,thick, dashed] (0.64, 0.45) rectangle (0.81,0.17);
                \draw[green, thick, dashed] (0.45, 0.65) rectangle (0.57, 0.43);
            \end{scope}
        \end{tikzpicture}
        \caption{InstantSplat \cite{fan2024instantsplat}}
    \end{subfigure}
    \caption{
        3DGS- or NeRF-update results:
        (a) Sparse RGB observations guiding NeRF or 3DGS update;
        (b) Evaluation-view renders by pre-trained 3DGS;
        (c) Ground truth evaluation images;
        (d-f) Evaluation-view renders by our method and baselines.
        The object \textcolor{green}{move-in} and \textcolor{red}{move-out} regions are highlighted with 
        \textcolor{green}{green} and \textcolor{red}{red} boxes respectively.
    }
    \label{fig:update}
\end{figure*}
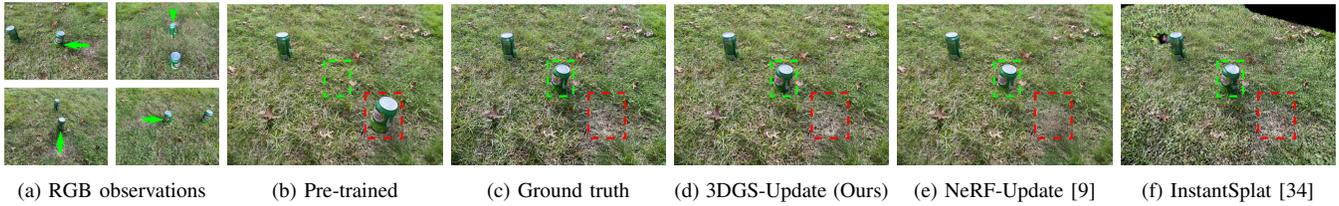

After training a 3DGS model for an initially static scene, if changes are made to the scene,
our method enables the update of the initial 3DGS to reflect physical changes with the guidance of \textit{sparse} post-change images.


Thanks to the explicit nature of the 3DGS, the Gaussians representing changed objects can be easily transformed to reflect object re-configurations.
A more challenging task, however, is reconstructing the previously un-observed regions in the 3DGS model that are revealed by the object rearrangement,
such as the part of the table that was previously occluded by the moved mustard bottle in Fig.~\ref{fig:qual}(a).
Instead of relying on in-painting-based solutions (e.g., GaussianEditor \cite{chen2024gaussianeditor}) to infer these missing parts, 
we directly leverage supervision from post-change images to fill in the gaps.

In our method, we transform the in-object pre-optimized Gaussians according to the estimated object movements,
and freeze all pre-trained Gaussian parameters.
We then duplicate the Gaussians near the newly un-occluded regions and optimize only the duplicates to fill in the missing parts.
Since these Gaussians are usually similar in position and color to the missing ones,
they provide a good initialization for subsequent 3DGS optimization under sparse-view supervision.
The Gaussians are optimized for 1000 step, during which 
we use the adaptive control strategy \cite{kerbl20233d} to split, duplicate, reset and cull Gaussians every 100 steps.

We test our method on the \textit{Soda} scene from the NeRF-Update dataset \cite{lu2024fast}, as shown in Fig.~\ref{fig:update}.
In this scene, 174 pre-change images are used to train a 3DGS and 4 post-change images serve as guidance for the update.
We evaluate our method on reconstruction quality for the soda can's move-in and move-out regions with PSNR and SSIM,
and also on runtime performance, as reported in Tab.~\ref{tab:update}.
Our method achieves comparable reconstruction quality and superior runtime performance compared to the NeRF-update~\cite{lu2024fast} method.
We also compare it with the SOTA sparse-view 3DGS training method, InstantSplat \cite{kerbl20233d},
which is only trained on the 4 post-change images.
While InstantSplat is faster, it delivers lower reconstruction quality.

\begin{table}[htb!]
\caption{
    \textbf{Quantitative evaluation} of 3DGS or NeRF update methods on the \textit{Soda} scene from NeRF-update dataset \cite{lu2024fast}.
    We report the average PSNR~($\uparrow$) and SSIM~($\uparrow$) for the move-in and move-out regions on evaluation images,
    and the 3DGS or NeRF update or InstantSplat \cite{fan2024instantsplat} training time.
}
\label{tab:update}
\centering
\begin{tabular}{c|cc|cc|c}
\hline
\multirow{2}{*}{Methods} & \multicolumn{2}{c|}{\textbf{Move-in}} 
& \multicolumn{2}{c|}{\textbf{Move-out}} & time \\
 & PSNR & SSIM & PSNR & SSIM & (s) \\
\hline
\textbf{3DGS-update} (\textbf{Ours}) & 17.01  & 0.46 
& \textbf{15.51} & \textbf{0.31} & 49 \\
NeRF-update & \textbf{17.39}  & \textbf{0.50} & 14.61 & 0.30 & 168 \\
InstantSplat & 15.35& 0.36  & 13.47 & 0.14 & \textbf{31} \\
\hline
\end{tabular}
\end{table}

\section{Limitations}
\textbf{Non-rigid object changes:}
Our method represents object pose changes as 6DoF rigid transformations,
which restricts its immediate application to non-rigid object changes.
However, our pipeline is designed to be modular.
Apart from the object pose estimation module, other components do not depend on the rigid object assumption.
This allows for easy integration of a non-rigid pose estimation method to accommodate such changes.

\textbf{Severe occlusions:}
Our method may fail if the changes are heavily occluded in all post-change views.
This could result in incomplete object templates and cause the object 3D segmentation to fail.
To mitigate this, we recommend capturing post-change images from angles that minimize occlusions, whenever possible.

\textbf{Foreground-background semantic similarity:}
The success of our 2D change detection module depends on the semantic distinction between the moved object and its background.
If the object and the background are semantically similar,
such as removing an apple from a pile of apples,
the change detection performance may degrade.
This issue can be mitigated by integrating our EfficientSAM-based approach with a photometric differencing method, 
allowing for the assessment of both semantic and visual differences between the foreground and background.

\section{Conclusions}
We develop a novel 3DGS-based change detection method for identifying 3D object-level changes in complex real-world environments.
Our approach significantly improves the accuracy and efficiency of radiance-field-based 3D change detection,
enabling a wide range of real-world applications.







\bibliographystyle{IEEEtran}
\bibliography{root}

\end{document}